\documentclass[11pt]{article}

\usepackage[final]{acl}

\usepackage{times}
\usepackage{latexsym}

\usepackage[T1]{fontenc}
\usepackage[utf8]{inputenc}

\usepackage{microtype}

\usepackage{inconsolata}

\usepackage{graphicx}

\usepackage{amssymb}
\usepackage{amsmath}
\usepackage{mathtools}
\usepackage{bbm}

\usepackage{algorithm} \usepackage{algpseudocode}
\usepackage{xurl}

\usepackage{booktabs}
\usepackage{multirow}
\usepackage{array}
\usepackage{xcolor}
\usepackage{colortbl}
\usepackage{graphicx}
\usepackage{stfloats}

\definecolor{headerBg}{HTML}{F8FAFC}      % 表头优雅浅灰（类Nature风格）
\definecolor{headerText}{HTML}{1E293B}    % 表头深灰文字（不用纯黑，更柔和）
\definecolor{categoryBg}{HTML}{E2E8F0}    % 分类行背景（稍深一点的灰）
\definecolor{categoryText}{HTML}{334155}  % 分类文字颜色
\definecolor{ourMethodBg}{HTML}{DCFCE7}   % 我们的方法（清新淡绿）
\definecolor{bestBg}{HTML}{DBEAFE}        % 最优结果背景（淡蓝色）
\definecolor{secondBg}{HTML}{FED7D7}      % 次优结果背景（柔和桃粉）
\definecolor{avgBestBg}{HTML}{FEF3C7}     % 平均值最优背景（暖金色-突出冠军）
\definecolor{avgSecondBg}{HTML}{E9D5FF}   % 平均值次优背景（柔和紫色-协调区分）
\definecolor{tokenColor}{HTML}{64748B}    % Token颜色（柔和灰蓝）
\definecolor{sectionBg}{HTML}{F1F5F9}     % 分组标题背景（极淡灰蓝）

\providecommand{\res}[3]{%
    #1\ifdim #2pt=0pt\else{\tiny$_{\pm #2}$}\fi \ {\textcolor{tokenColor}{\scriptsize (#3)}}%
}
\providecommand{\best}[3]{%
    \cellcolor{bestBg}\textbf{#1}\ifdim #2pt=0pt\else{\tiny$_{\pm #2}$}\fi \ {\textcolor{tokenColor}{\scriptsize (#3)}}%
}
\providecommand{\second}[3]{%
    \cellcolor{secondBg}#1\ifdim #2pt=0pt\else{\tiny$_{\pm #2}$}\fi \ {\textcolor{tokenColor}{\scriptsize (#3)}}%
}

\providecommand{\avgbest}[2]{%
    \cellcolor{avgBestBg}\textbf{#1} \ {\textcolor{tokenColor}{\scriptsize (#2)}}%
}
\providecommand{\avgsecond}[2]{%
    \cellcolor{avgSecondBg}#1 \ {\textcolor{tokenColor}{\scriptsize (#2)}}%
}
\providecommand{\avg}[2]{%
    #1 \ {\textcolor{tokenColor}{\scriptsize (#2)}}%
}

\usepackage[most]{tcolorbox}
\usepackage{xcolor}

\definecolor{failFrame}{RGB}{200, 60, 60}    % 深红边框
\definecolor{failBack}{RGB}{255, 245, 245}   % 极淡红背景
\definecolor{succFrame}{RGB}{40, 160, 40}    % 深绿边框
\definecolor{succBack}{RGB}{245, 255, 245}   % 极淡绿背景
\definecolor{mainFrame}{RGB}{80, 80, 80}     % 主盒子深灰边框

\newcommand{\txtspiral}[1]{\textcolor{failFrame}{\textbf{\textit{#1}}}} % 红色螺旋高亮
\newcommand{\txtcorrect}[1]{\textcolor{succFrame}{\textbf{#1}}}         % 绿色修正高亮

\title{Dissecting Failure Dynamics in Large Language Model Reasoning}

\author{
    Wei Zhu, Jian Zhang, Lixing Yu, Kun Yue, Zhiwen Tang\thanks{Corresponding author.} \\
    School of Information Science and Engineering, Yunnan University, Kunming, China \\
    Yunnan Key Laboratory of Intelligent Systems and Computing, Kunming, China \\
    \texttt{zhuwei@stu.ynu.edu.cn, zhiwen.tang@ynu.edu.cn}
}

\begin{document}
\maketitle
\begin{abstract}
% This document is a supplement to the general instructions for *ACL authors. It contains instructions for using the \LaTeX{} style files for ACL conferences.
% The document itself conforms to its own specifications, and is therefore an example of what your manuscript should look like.
% These instructions should be used both for papers submitted for review and for final versions of accepted papers.

% Reasoning failures in large language models (LLMs) are commonly addressed by allocating more computation at inference time. In this work, we show that this assumption is misaligned with how reasoning actually fails. Through a systematic analysis of reasoning trajectories, we find that failures follow a structured dynamic. Errors typically originate from a sparse early divergence, after which the model continues to produce locally coherent but globally invalid reasoning. Importantly, many failures do not stem from missing knowledge, but from a loss of navigational stability during inference. We further show that these structural fractures are accompanied by localized spikes in model uncertainty, which serve as diagnostic signals of instability rather than objectives for global search. 

% Guided by these findings, we introduce a minimal test-time intervention that performs targeted, in-place steering on a single reasoning trajectory when instability is detected. Experiments across diverse reasoning benchmarks demonstrate consistent accuracy improvements under fixed computational budgets, with substantially reduced redundancy compared to conventional test-time scaling methods.

Large Language Models (LLMs) achieve strong performance through extended inference-time deliberation, yet how their reasoning failures arise remains poorly understood. By analyzing model-generated reasoning trajectories, we find that errors are not uniformly distributed but often originate from a small number of early transition points, after which reasoning remains locally coherent but globally incorrect. These transitions coincide with localized spikes in token-level entropy, and alternative continuations from the same intermediate state can still lead to correct solutions. Based on these observations, we introduce GUARD\footnote{Code is available at \url{https://github.com/ZHUWEI-hub/GUARD}.}, a targeted inference-time framework that probes and redirects critical transitions using uncertainty signals. Empirical evaluations across multiple benchmarks confirm that interventions guided by these failure dynamics lead to more reliable reasoning outcomes. Our findings highlight the importance of understanding when and how reasoning first deviates, complementing existing approaches that focus on scaling inference-time computation.

\end{abstract}

% \section{Introduction}
\section{Introduction}

% Large Reasoning Models (LRMs), such as OpenAI o1 \citep{jaech2024openai} and DeepSeek-R1 \citep{model-deepseek-r1}, approximate human-like "System 2" reasoning by internalizing test-time scaling—generating extended chains of thought to decompose complex problems. \cite{wei2022chaincot, yao2023tot, GOT} This reinforcement learning-driven deliberation allows models to self-correct and handle tasks requiring high-order cognitive processing.

% Despite these capabilities, LRMs exhibit recurring failure patterns across diverse benchmarks. A common response involves allocating additional inference compute via longer chains \citep{muennighoff-etal-2025-s1}, parallel sampling \citep{cot-sc2023, scalena2025eager, xu2025dts}, or process optimization strategies \citep{zhang-etal-2025-entroduction, zhang-etal-2025-alphaone, snell2025scaling}. These approaches implicitly assume that errors stem from insufficient deliberation or sub-optimal efficiency. However, by prioritizing outcome rectification over process diagnosis, they overlook the specific moments where generation first diverges. Consequently, even optimized strategies incur computational costs that are disproportionate to the often localized nature of the error.

\begin{figure}[!t]
  \centering
    \includegraphics[width=\columnwidth]{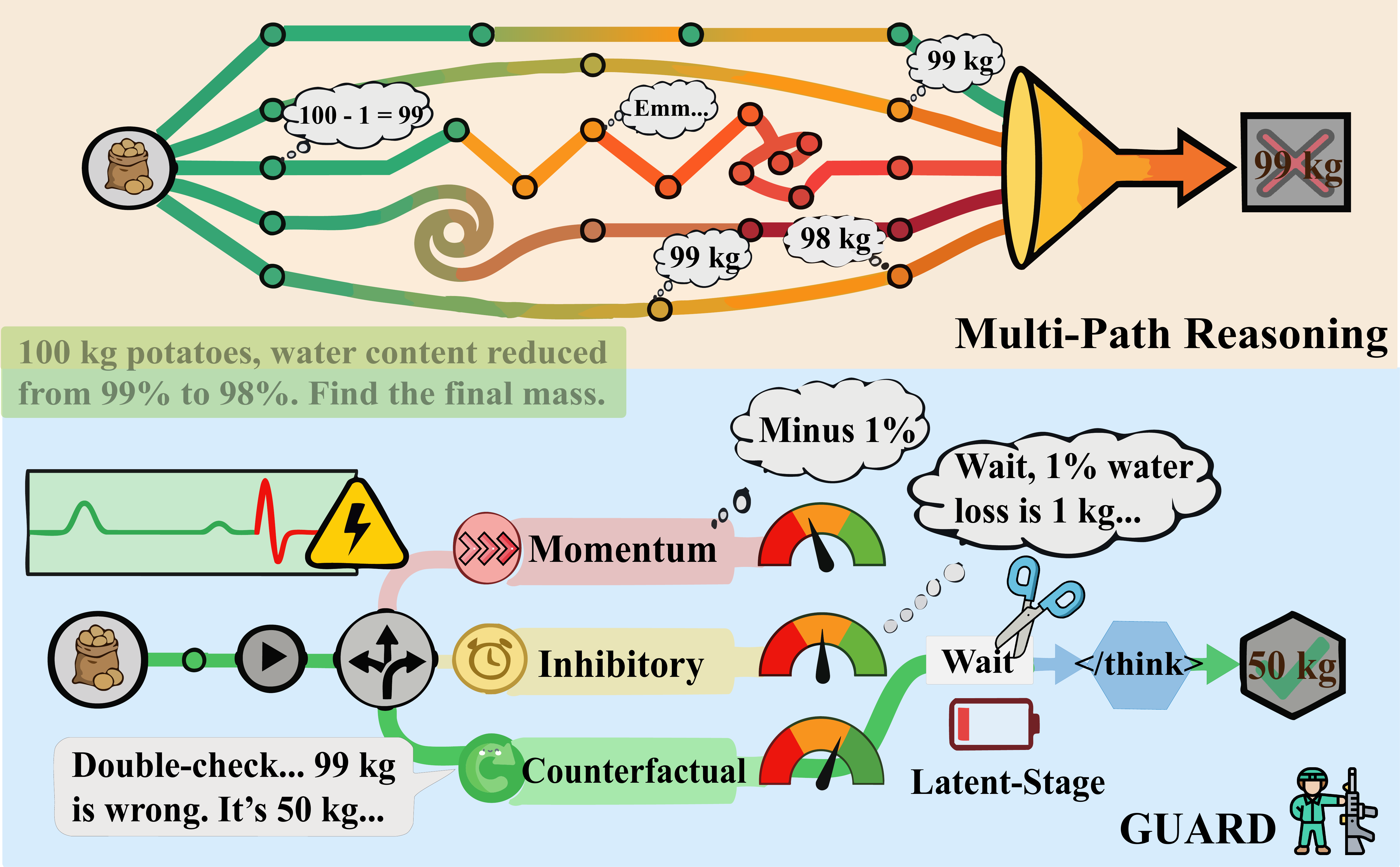}
  \caption{\textbf{Comparison of Multi-path Reasoning versus GUARD.} Multi-path reasoning relies on repeated sampling of parallel trajectories, whereas GUARD maintains a single primary trajectory and intervenes only at critical transitions using targeted branching.}
  \label{fig:temporal_distribution}
\end{figure}

Large Reasoning Models (LRMs), such as OpenAI o1 \citep{jaech2024openai} and DeepSeek-R1 \citep{model-deepseek-r1}, aim to approximate human-like deliberative reasoning by internalizing test-time scaling. Through extended chains of thought, these models decompose complex problems into intermediate steps, enabling multi-stage reasoning and iterative refinement \citep{wei2022chaincot, yao2023tot, GOT}. Reinforcement learning further strengthens this capability by encouraging sustained deliberation on challenging tasks \citep{uesato2022solving-prm,lightman2023let-prm, he2025justrl, guo2025thinkingprogrammingvisionunified, hong2025generativereasoningrecommendationllms, ma2026tspo, wu2026steppotentialadvantageestimation}.

Consequently, much recent progress has focused on allocating additional inference-time computation to improve reasoning performance. Representative approaches include generating longer reasoning traces \citep{liu2023retrieval, snell2025scaling, muennighoff-etal-2025-s1, li2026retrievalgenerationunifiedframework}, sampling multiple trajectories in parallel \citep{cot-sc2023, snell2025scaling, scalena2025eager, xu2025dts}, and optimizing inference-time procedures \citep{zhang-etal-2025-entroduction, zhang-etal-2025-alphaone}. These methods have demonstrated clear gains across benchmarks, reinforcing the view that increased deliberation can be beneficial. 
Yet these gains provide limited insight into where reasoning  goes wrong within a single trajectory,  and whether such deviations are isolated events or systematically concentrated in time.

In this work, we address this question by analyzing reasoning failures at the trajectory level. Rather than treating incorrect outputs as undifferentiated outcomes, we examine how errors emerge and evolve over time within a single reasoning trace. By systematically analyzing model-generated reasoning trajectories, we study when failures first occur, how they affect subsequent steps, and whether their influence is evenly spread or temporally concentrated.

Our analysis uncovers clear regularities in failure dynamics. Reasoning errors are often temporally concentrated, with failure onsets occurring disproportionately early in the trajectory. After such an onset, the model typically continues with locally coherent but globally incorrect reasoning, allowing early deviations to exert a lasting downstream influence. These critical transitions are marked by localized spikes in token-level entropy, while uncertainty elsewhere remains stable. Moreover, alternative continuations from the same intermediate state can still reach correct solutions, indicating that many failures arise from specific transition choices rather than missing task-relevant knowledge.

Guided by these findings, we introduce \textbf{G}uided \textbf{U}ncertainty-\textbf{A}ware \textbf{R}easoning with \textbf{D}ecision control (\textbf{GUARD}), a lightweight inference-time framework for correcting reasoning trajectories. Rather than expanding computation globally or maintaining multiple parallel paths throughout generation, GUARD follows a single primary reasoning trajectory and introduces only short-horizon local branching when high-risk transitions are detected. These brief interventions allow the model to reconsider critical steps while continuing generation along a single evolving solution. By steering generation away from early deviations and suppressing unproductive late-stage expansion, GUARD improves reasoning reliability without altering the underlying model.

The remainder of this paper is organized as follows. Section~2 reviews related work. Section~3 analyzes recurring failure dynamics in LRM reasoning. Section~4 presents the GUARD framework and its intervention mechanisms. Section~5 evaluates GUARD across multiple benchmarks and model backbones.

\section{Related Work}

\subsection{Large Reasoning Models}
% The paradigm of LLM reasoning has shifted from prompt-induced behaviors to intrinsic architectural capabilities. While early Chain-of-Thought (CoT) methods~\cite{wei2022chain} relied on few-shot prompting to elicit intermediate steps, recent \textbf{Large Reasoning Models (LRMs)} internalize this process to achieve System 2-like deliberation~\cite{kahneman2011thinking}. Models such as OpenAI's o1~\cite{jaech2024} and DeepSeek-R1~\cite{deepseek2025} utilize special control tokens, such as \texttt{<think>}, to demarcate extended latent thought trajectories. These internal reasoning paths are optimized via large-scale reinforcement learning (RL) to align with verifiable outcomes~\cite{uesato2022solving}. Furthermore, the emergence of open-weights models and distillation techniques~\cite{deepseek2025, qwen2025} has democratized access to these dense reasoning traces. However, simply scaling chain length does not guarantee correctness. Consequently, extended reasoning often amplifies redundancy rather than accuracy, leading to computational waste. This necessitates mechanisms to dynamically modulate inference.
The reasoning landscape has shifted from prompt-induced CoT~\cite{cot-sc2023} to intrinsic Lar ge Reasoning Models (LRMs). Models like OpenAI's o1~\cite{jaech2024openai} and DeepSeek-R1~\cite{model-deepseek-r1} internalize System 2 deliberation, employing latent trajectories optimized via reinforcement learning~\cite{uesato2022solving-prm,lightman2023let-prm, he2025justrl}. While these reasoning-centric models ~\cite{model-deepseek-r1,qwen_team_qwq_2025,team2024qwen2,abdin2024phi, liu2025queries} demonstrate strong deliberative capabilities, explicit long-form reasoning remains an inefficient proxy for deliberation, as longer chains frequently increase redundancy without accuracy gains, motivating dynamic inference regulation.

% \subsection{Test-Time Scaling Strategies}
% \label{subsec:test_time_scaling}

% Test-time scaling strategies empower LLMs to trade inference compute for performance, utilizing paradigms ranging from sequential refinement~\cite{reflexion} to parallel sampling (e.g., Best-of-N, Tree-of-Thoughts) and Monte Carlo Tree Search (MCTS)~\cite{wang2023self, xie2023decomposition, zhou2025}. While effective, mainstream structured search methods often rely on blind scaling or expensive verifiers (PRMs)~\cite{lightman2023let}, resulting in significant computational redundancy. To mitigate this inefficiency, a growing body of work leverages intrinsic uncertainty to modulate the search process. For instance, DTS~\cite{xu2025} triggers selective branching based on absolute entropy values, whereas EGB~\cite{li2025} combines entropy gating with PRMs. Similarly, EAGER~\cite{scalena2025} and Entroduction~\cite{zhang2025} dynamically reallocate budget based on uncertainty signals. However, these frameworks predominantly rely on static, pre-defined thresholds that lack adaptability across varying task distributions, or necessitate computationally expensive external verifiers. Crucially, they typically default to generating multiple complete parallel solutions to ensure coverage. In contrast, our approach utilizes an adaptive threshold derived from the model's own historical entropy percentile, performing low-budget, in-place interventions to rectify a \textit{single} reasoning trajectory without the overhead of maintaining concurrent full-length hypotheses.

\subsection{Test-Time Scaling Strategies}
\label{subsec:test_time_scaling}

Test-time scaling empowers LLMs to trade inference compute for performance via paradigms ranging from sequential refinement~\cite{shinn2023reflexion, snell2025scaling} to parallel sampling (e.g., Best-of-N, Tree-of-Thoughts) and Monte Carlo Tree Search (MCTS)~\cite{cot-sc2023, yao2023tot, symphony}. However, mainstream methods often rely on blind scaling or expensive verifiers ~\cite{wang2025pats,liao2025reward}, incurring significant redundancy. To mitigate this, recent works leverage intrinsic uncertainty. DTS~\cite{xu2025dts} triggers selective branching based on absolute entropy, while EGB~\cite{li2025EGB} combines entropy gating with PRMs. Similarly, EAGER~\cite{scalena2025eager} and Entro-duction~\cite{zhang-etal-2025-entroduction} dynamically reallocate budgets. Yet, these frameworks typically rely on static, non-adaptive thresholds or external verifiers. Crucially, they often generate multiple complete parallel solutions. In contrast, our approach uses an adaptive threshold based on historical entropy percentiles, performing low-budget, in-place interventions on a \textit{single} trajectory without maintaining concurrent hypotheses.

\subsection{Efficient Reasoning}
\label{subsec:efficient_reasoning}

Parallel to scaling, extensive work has examined inference efficiency. Chain of Draft ~\cite{xu2025cod} was introduced to enforce minimalism, though often at the cost of zero-shot accuracy. Collaborative frameworks~\cite{liao2025reward,chen2025R-stich,fu2025r2r,yang2025speculative, lian2026sweagilesoftwareagentframework, nie2026attnpo} offload steps to lighter models but incur alignment complexity and switching overheads. Dynamic strategies like CGRS~\cite{huang2025CGRS}, DEER~\cite{yang2025DEER}, Adaptive Think~\cite{yong2025thinkprnot}, and $\alpha1$~\cite{zhang-etal-2025-alphaone} modulate depth via confidence or information-theoretic metrics, yet suffer from rigid heuristics or dependencies on pre-computed statistics. Fundamentally, these paradigms prioritize minimizing length, ignoring the algorithmic overhead of control mechanisms and the mining of latent capabilities. In contrast, our approach targets capability maximization with minimal redundancy, repairing fractures rather than merely shortening trajectories.

\section{Empirical Findings on Reasoning Failure Dynamics}
\label{sec:empirical_analysis}
In this section, we analyze how reasoning failures arise and propagate along a single generated trajectory. By examining model-produced reasoning traces, we observe several recurring characteristics in how failures develop along the trajectory. Errors often emerge early, expand through subsequent locally coherent steps, exhibit localized uncertainty signatures, and are sometimes recoverable from the same intermediate state. These findings provide a trajectory-level characterization of reasoning failure.

% Our analysis is conducted on reasoning trajectories generated by DeepSeek-R1-Distill-Qwen-1.5B~\cite{model-deepseek-r1} on AMC \cite{amc23} and AIME~\cite{aime} benchmarks. Each output is decomposed into an ordered sequence of reasoning segments 
% $\tau = (s_k)_{k=1}^K$, where segments correspond to semantically coherent steps separated by the delimiter \verb|\n\n|. Segment-level validity is assessed using Gemini 3 Pro \cite{google2025gemini3} and human experts as external verifiers, referred to as the \textit{Oracle}. A segment is marked invalid if it introduces an incorrect assumption or deduction that precludes reaching the correct final answer.

Our analysis is based on reasoning trajectories generated by DeepSeek-R1-Distill-Qwen-1.5B~\cite{model-deepseek-r1} on the AMC~\cite{amc23} and AIME~\cite{aime} benchmarks. Each output is segmented into an ordered sequence $\tau = (s_k)_{k=1}^K$ using the delimiter \verb|\n\n|. Segment validity is evaluated using an external oracle based on Gemini 3 Pro~\cite{google2025gemini3}, with human verification for quality control. A segment is labeled invalid if it introduces an error that prevents reaching the correct final answer.

\subsection{Early Failure Onsets}
\label{subsec:temporal}
\begin{figure}[t]
  \centering
  \includegraphics[width=\columnwidth]{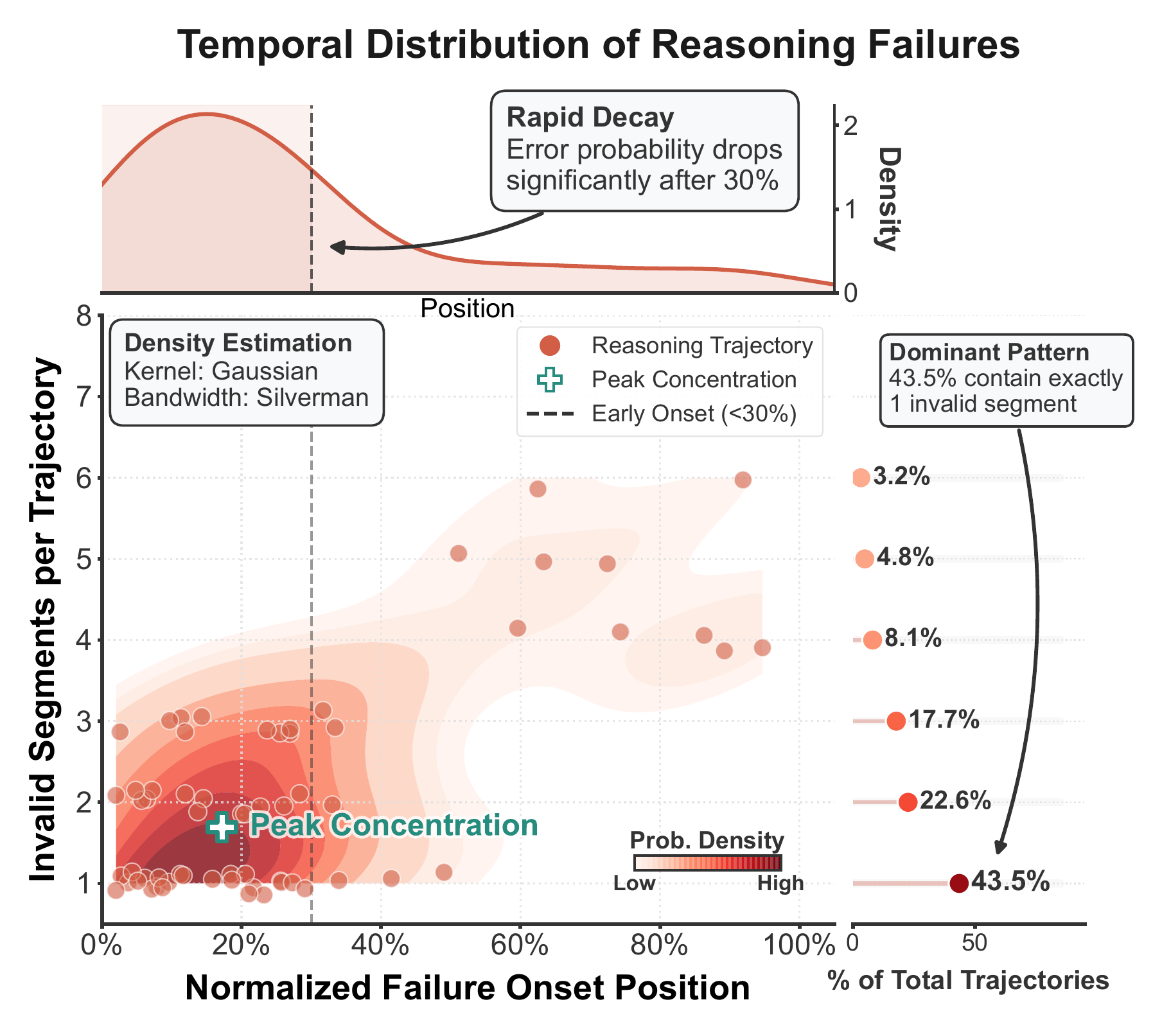}
  \caption{\textbf{Early Concentration of Reasoning Failures.} Failure onsets are heavily concentrated in the early stages of generation, and most incorrect trajectories contain only a small number of invalid segments, with 43.5\% exhibiting a single error.}
  \label{fig:temporal_distribution}
\end{figure}

We begin by examining when reasoning failures arise along a generated trajectory. For each reasoning trace $\tau = (s_k)_{k=1}^K$, we use the Oracle to assign a segment-level validity label $\mathcal{O}(s_k) \in \{0,1\}$ where $\mathcal{O}(s_k) =1$ indicates that segment $s_k$ is logically valid with respect to the problem context and preceding segments, and $\mathcal{O}(s_k) =0$ otherwise. We define a \textit{failure onset} at segment $s_k$ whenever $\mathcal{O}(s_{k-1}) = 1 \land \mathcal{O}(s_k) = 0$. This definition captures the transition from a valid reasoning prefix to an invalid step.  

Figure~\ref{fig:temporal_distribution} visualizes the temporal distribution of failure onsets. The top panel shows a strong early concentration, with over 85\% of failure onsets occurring within the first 30\% of the trajectory. The bottom panel presents the joint distribution of normalized failure onset position and the number of invalid segments per trajectory, estimated using a Gaussian kernel with Silverman bandwidth. The density exhibits a dominant concentration corresponding to early-stage failures accompanied by one to two invalid segments. In particular, 43.5\% of trajectories contain exactly one invalid segment. 

These patterns indicate that reasoning failures are typically driven by early, localized deviations that account for most errors within a trajectory, rather than by difficulty that accumulates uniformly over time. The concentration of failure onsets in a small number of early segments suggests that the downstream behavior of a trajectory is often determined by a limited set of critical transitions, highlighting the importance of identifying such moments during generation.

\subsection{Post-Onset Trajectory Expansion}
\label{subsec:length}
\begin{figure}[t]
  \centering
  \includegraphics[width=\columnwidth]{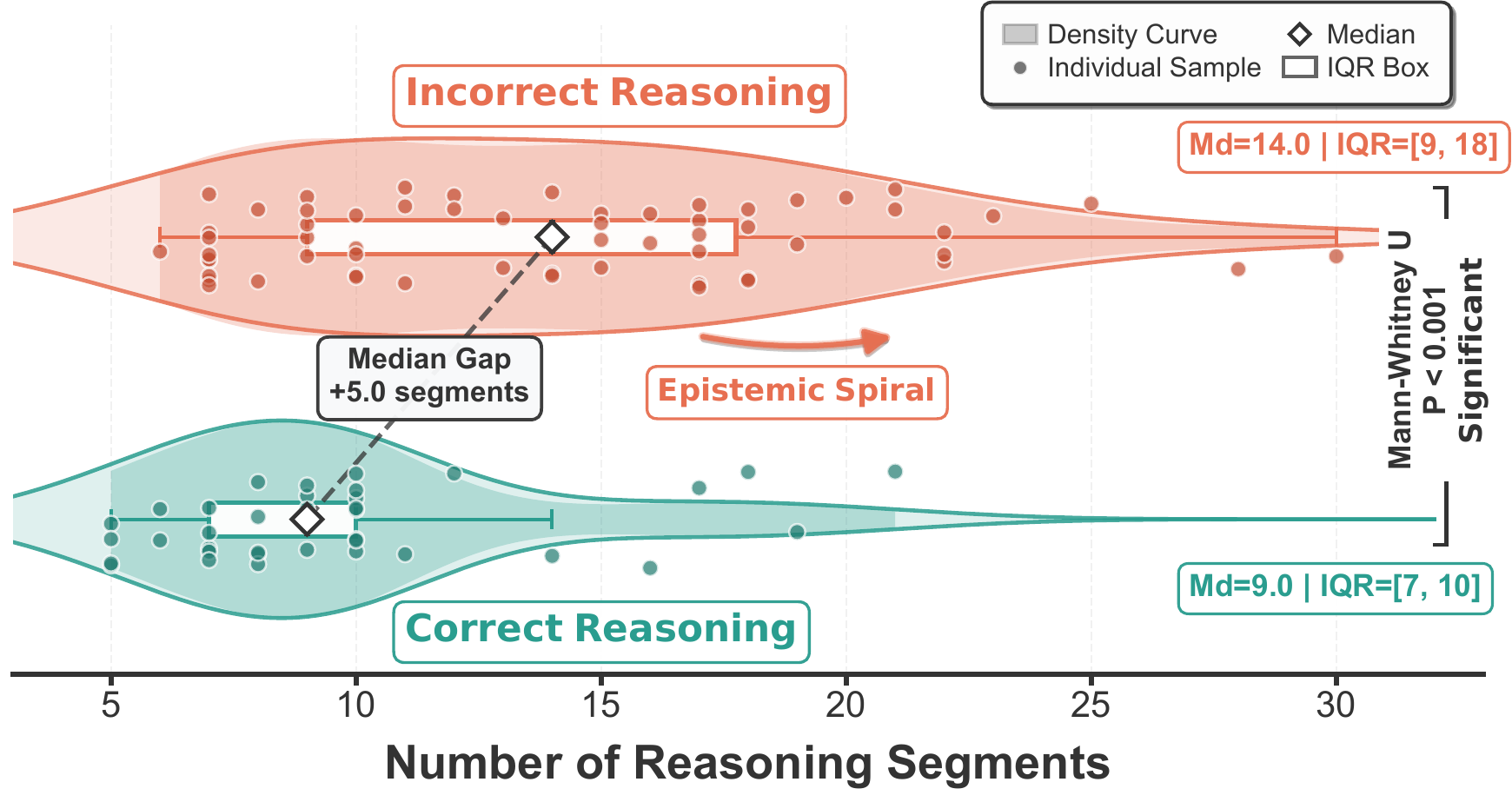}
  \caption{\textbf{Segment Count Distribution for Correct and Incorrect Trajectories.} Incorrect trajectories exhibit substantial length expansion following failure onsets.%, forming an epistemic spiral where additional reasoning steps elaborate and reinforce an initial incorrect premise rather than correcting it.
  }
  \label{fig:number_of_segment_distribution}
\end{figure}

\begin{figure*}[t]
  \centering
  \begin{minipage}{0.48\linewidth}
    \centering
    \includegraphics[width=\linewidth]{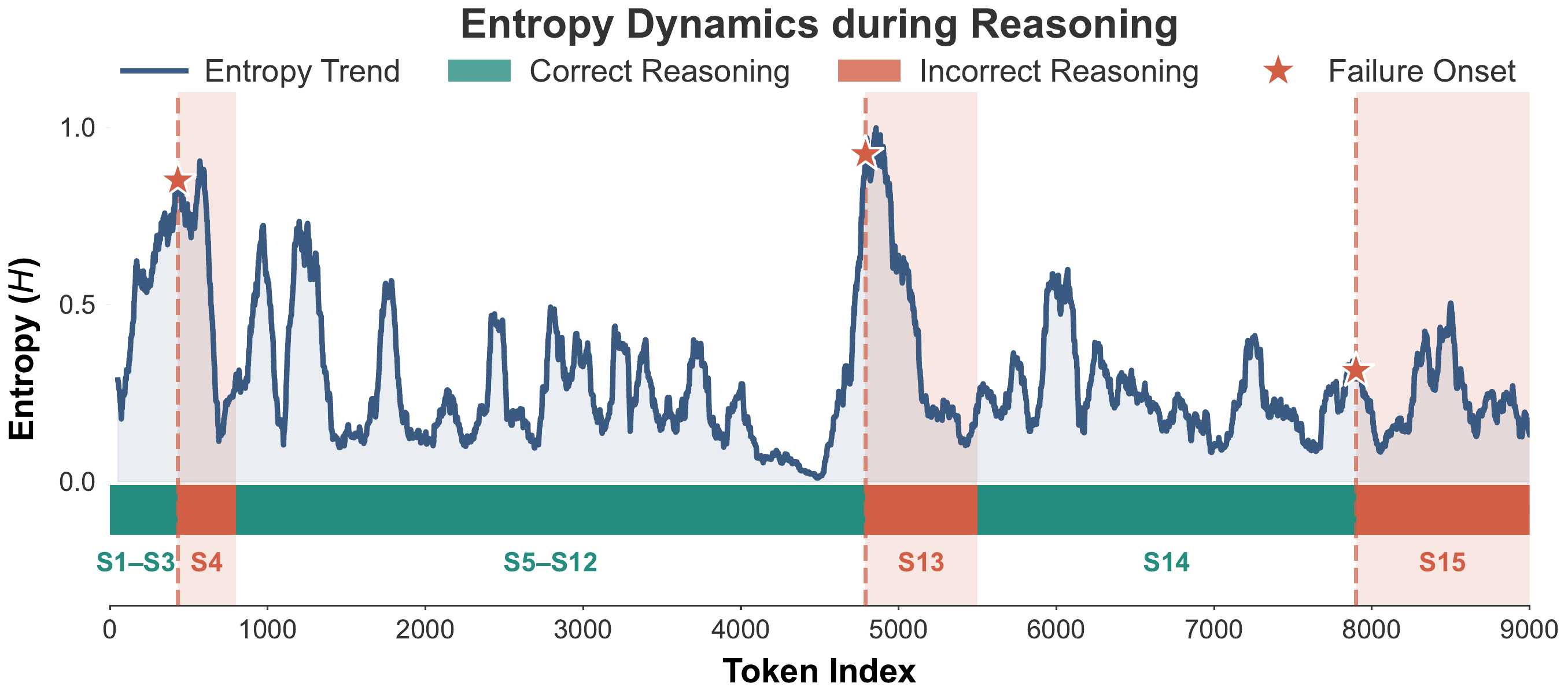}
  \end{minipage}
  \hfill
  \begin{minipage}{0.48\linewidth}
    \centering
    \includegraphics[width=\linewidth]{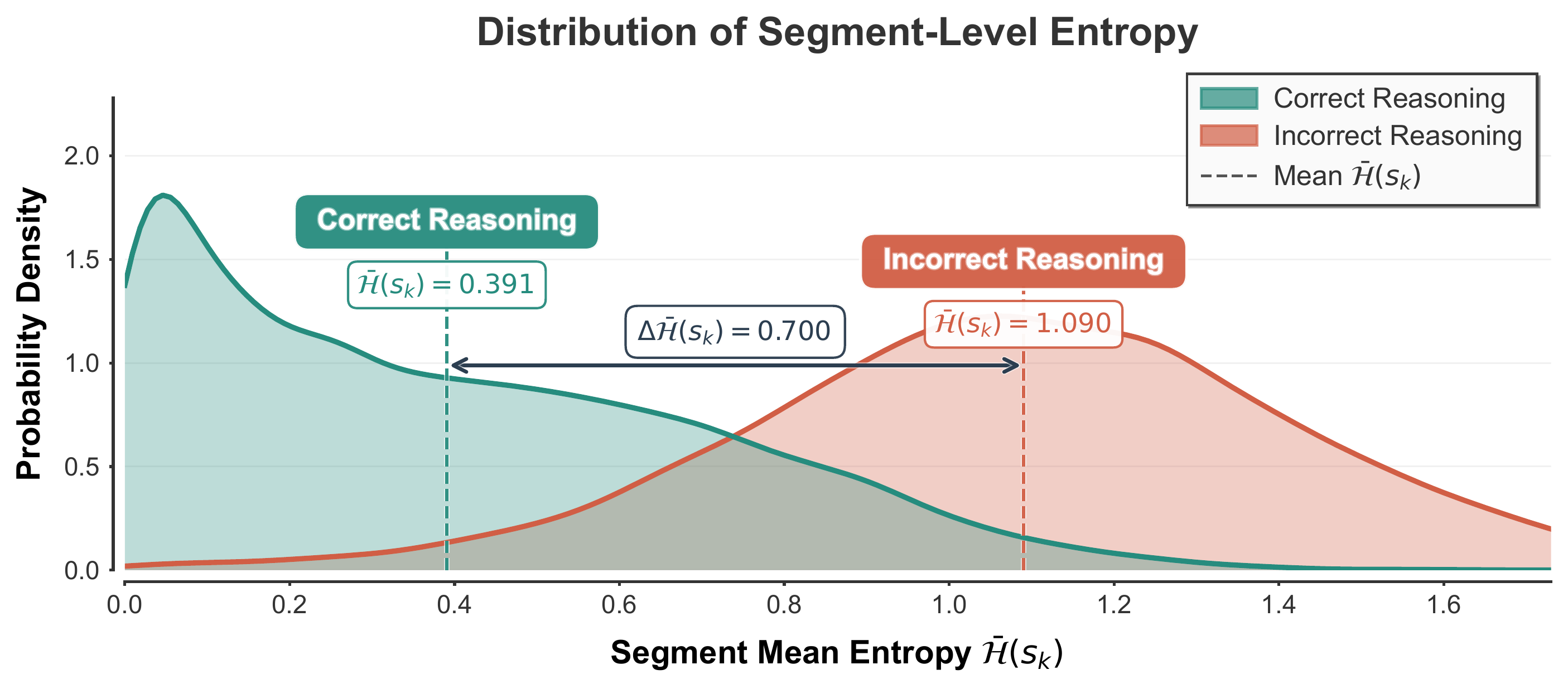}
  \end{minipage}
  \caption{ \textbf{Left: Entropy aligned to failure onset}, with a localized spike at the transition to invalid reasoning. \textbf{Right: Entropy density for valid and invalid segments}, showing higher dispersion and a shifted mean for error segments.}
 \label{fig:entropy_distribution}
\end{figure*}

We next examine how the length of a reasoning trajectory relates to its correctness. As shown in Figure~\ref{fig:number_of_segment_distribution}, incorrect trajectories contain substantially more reasoning segments than correct ones, exhibiting a pronounced long tail in the segment-count distribution.

This length expansion occurs predominantly after the failure onset. Section \ref{subsec:temporal} shows that most failure onsets arise early in the trajectory, whereas incorrect trajectories continue to generate many additional segments thereafter. Notably, these post-onset segments are not syntactically degenerate or abruptly incoherent. Instead, they form extended sequences of locally plausible reasoning that remain consistent with the initial erroneous premise. We refer to this empirical pattern as an \textit{epistemic spiral}, characterizing the sustained expansion of reasoning following an early failure. Examples of epistemic spiral can be found in Appendix~\ref{app:case_study}.

As a result, trajectory length is dominated by post-onset expansion, and extended reasoning is strongly associated with incorrect outcomes, suggesting limited benefit from allocating additional computation to long trajectories.

% We next compare the overall length of correct and incorrect reasoning trajectories. Figure~\ref{fig:number_of_segment_distribution} shows that incorrect trajectories contain significantly more reasoning segments than correct ones. Incorrect trajectories exhibit a pronounced long tail in the segment count distribution.

% Importantly, this expansion occurs predominantly after the failure onset. As shown in Figure \ref{fig:temporal_distribution}, most failure onsets arise early in the trajectory, yet the generation process continues for many additional segments. After the failure happens, the model does not immediately collapse into incoherent reasoning. Instead, it enters what we term an \textit{epistemic spiral}, repeatedly generating locally plausible justifications that rationalize the initial erroneous assumption. These segments remain internally consistent with the corrupted premise while drifting further away from the correct solution.

% % 此段缺乏根据，只是对现象的一种解释
% This behavior explains why incorrect trajectories tend to be longer despite early failure. Once failure occurs, the model allocates additional reasoning steps to elaborate and defend an incorrect line of thought, leading to redundant and self-reinforcing expansions. As a result, extending the reasoning process amplifies the consequences of early divergence rather than correcting it.

\subsection{Elevated Uncertainty in Error Segments}
\label{subsec:uncertainty}

We next examine whether reasoning errors are accompanied by systematic changes in model uncertainty. Let $\mathbf{z}_t \in \mathbb{R}^{|\mathcal{V}|}$ denote the model logits at token position $t$. The next-token probability distribution $P$ is defined via the softmax transformation:
\begin{equation}
    P(x_t = v \mid x_{<t}) = \frac{\exp(\mathbf{z}_{t}[v])}{\sum_{v' \in \mathcal{V}} \exp(\mathbf{z}_{t}[v'])}.
\end{equation}
We quantify uncertainty using token-wise Shannon entropy $\mathcal{H}(x_t \mid x_{<t})$ and its length-normalized segment aggregation.  For a reasoning segment $s_k$ spanning a token subsequence $x_{u_k}, x_{u_k+1}, \dots , x_{v_k} $, we define:
\begin{equation}
\begin{split}
    \mathcal{H}(x_t \mid x_{<t}) &\coloneqq - \mathbb{E}_{v \sim P(\cdot \mid x_{<t})} \left[ \log P(v \mid x_{<t}) \right], \\
    \bar{\mathcal{H}}(s_k) &\coloneqq \frac{1}{|s_k|} \sum_{t=u_k}^{v_k} \mathcal{H}(x_t \mid x_{<t}).
\end{split}
\end{equation}
This normalization removes segment-length effects, enabling direct comparison of uncertainty across segments. 

We relate these uncertainty measures to the failure onset positions. Figure~\ref{fig:entropy_distribution} (left) shows pronounced \emph{local entropy spikes} at failure onsets, where segments corresponding to the onset exhibit a sharp increase in $\bar{\mathcal{H}}(s_k)$ relative to nearby segments. Figure~\ref{fig:entropy_distribution} (right) further shows a \emph{global entropy increase} for invalid segments compared to valid ones. Valid segments concentrate in a low-entropy regime, whereas invalid segments form a long-tailed distribution with a significantly higher mean uncertainty ($p<0.001$). 

These results show that uncertainty changes are tightly coupled to where errors arise. Elevated segment entropy marks brief transitions associated with the onset of failure and remains higher in subsequent invalid segments, providing a consistent signal that distinguishes erroneous reasoning from valid progression.

\subsection{Local Recoverability of Failures}
\label{subsec:recoverability}

\begin{figure}[t]
  \centering
  \includegraphics[width=\columnwidth]{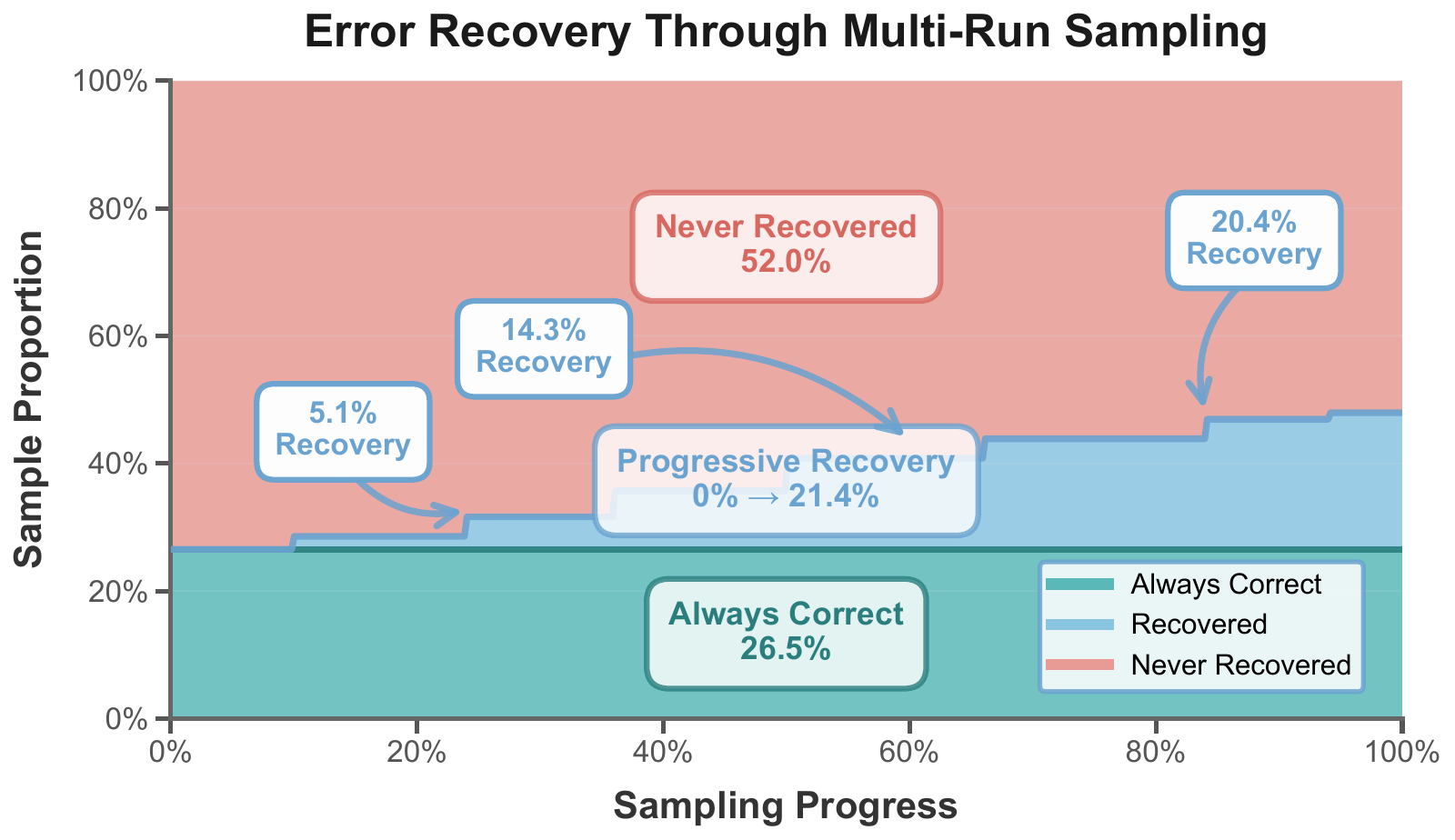}
  \caption{\textbf{Recoverability of Reasoning Failures.} Some failures persist across continuations, while others admit correct solutions from the same prefix.}
  \label{fig:recoverability_analysis}
\end{figure}

We next examine whether reasoning failures reflect irreversible loss or arise from recoverable trajectory choices. To this end, we analyze alternative continuations from the same intermediate state around each failure onset.

% For a reasoning trajectory $\tau$ with a failure onset at segment $s_k$, we treat the last valid segment, $s_{k-1}$, as an anchor. From the corresponding prefix, we generate multiple alternative continuations via stochastic sampling while keeping the prompt, context, and model parameters unchanged. A failed trajectory is considered \emph{locally recoverable} if at least one alternative continuation from the same intermediate state, meaning an identical reasoning prefix, reaches a correct final answer. This definition restricts recovery to variations in the continuation alone, without modifying earlier reasoning steps or introducing external information.

For a reasoning trajectory $\tau$ with a failure onset at segment $s_k$, we treat the last valid segment, $s_{k-1}$, as an anchor and generate multiple alternative continuations from the corresponding prefix via stochastic sampling.  A failed trajectory is considered \emph{locally recoverable} if at least one alternative continuation from this prefix reaches a correct final answer. This definition focuses on variability in continuation from the same valid prefix, without introducing additional information.

Figure~\ref{fig:recoverability_analysis} shows that more than 20\%  trajectories satisfy this criterion. In these cases, correct solutions remain reachable from the same prefix despite failure in the original trajectory, indicating that the error arises from the specific continuation taken after the onset rather than from an absence of viable reasoning paths. Recoverable cases therefore constitute a substantial subset of failures rather than isolated exceptions.

These observations indicate that early failures do not uniquely determine reasoning outcomes. Even when a trajectory diverges and subsequently expands through erroneous reasoning, alternative continuations from the same prefix can still reach correct solutions, highlighting the role of trajectory choice in shaping reasoning behavior.

\section{Guided Uncertainty-Aware Inference Control}
Motivated by the observed failure dynamics, we propose \textbf{G}uided \textbf{U}ncertainty-\textbf{A}ware \textbf{R}easoning with \textbf{D}ecision control (\textbf{GUARD}), a lightweight test-time approach for intervening during LLM reasoning. GUARD monitors uncertainty signals computed from the model’s next-token distribution and triggers intervention only at moments indicative of imminent failure. When triggered, it performs short-horizon branching to obtain a small set of candidate continuations and then selects a continuation based on entropy reduction, avoiding extensive search. In addition, GUARD incorporates a lightweight control mechanism for late-stage reasoning, where prolonged trajectory expansion is unlikely to yield correction. The remainder of this section describes the uncertainty signals used for triggering, the branch-and-select procedure, and the late-stage control mechanism.

\subsection{Detecting Failure Onsets}
\label{subsec:detection}

Elevated entropy often coincides with critical transitions that precede reasoning errors, making uncertainty a useful signal for selective intervention. We therefore monitor the token-wise Shannon entropy $\mathcal{H}(x_t)$ during generation and detect atypical spikes relative to the uncertainty observed so far.

To avoid brittle absolute thresholds, we compare the instantaneous entropy to an instance-adaptive baseline defined by a quantile of the entropy history $\mathbf{H}_{<t}$. This relative criterion identifies sharp uncertainty increases under the current prefix while remaining insensitive to the overall entropy scale.

Intervention is evaluated only at reasoning-step boundaries, where a new segment begins and local modifications can be applied without interrupting an ongoing step. Let $\mathcal{T}_{\text{delim}}$ denote the set of delimiter tokens (e.g., \verb|\n\n|). For the token $x_t$ immediately following such a delimiter, we define

% Our empirical analysis in Section~\ref{sec:empirical_analysis} reveals a tight coupling between uncertainty and reasoning errors. Elevated entropy often coincides with critical transition points in the reasoning process, serving as a reliable signal for identifying high-risk moments.

% We operationalize this using the token-wise Shannon entropy $\mathcal{H}(x_t)$. Let $\mathbf{H}_{<t}$ denote the history of observed token entropies within the same generation. Rather than relying on an absolute threshold, we compare the instantaneous entropy to an instance-adaptive baseline defined by a high quantile of $\mathbf{H}_{<t}$.

% In practice, we restrict monitoring to the onset of new reasoning segments. Let $\mathcal{T}_{\text{delim}}$ denote a set of structural delimiters (e.g., \verb|\n\n| and its variants). We evaluate the trigger condition on the token $x_t$ \textbf{immediately following} such a boundary:
\begin{equation}
\begin{split}
    \mathbb{I}_{\text{drift}}(x_t) = \mathbb{I}\big[ & x_{t-1} \in \mathcal{T}_{\text{delim}} \;\land \\
    & \mathcal{H}(x_t) > \text{Quantile}_q(\mathbf{H}_{<t}) \big],
\end{split}
\end{equation}
where $q \in (0,1)$ controls the sensitivity of the detector. When $\mathbb{I}_{\text{drift}}(x_t)=1$, GUARD activates the short-horizon branching procedure described in Section~\ref{subsec:branch}. This detection mechanism restricts intervention to a small number of high-risk transitions, avoiding unnecessary interference during routine generation.

\subsection{Branching at Failure Onsets}
\label{subsec:branch}
% Section~\ref{sec:empirical_analysis} shows that after a failure onset, erroneous trajectories often continue with locally coherent but globally incorrect reasoning, while alternative continuations from the same prefix may still lead to correct outcomes. This motivates a localized intervention that briefly explores a small set of alternative continuations immediately after detection, rather than diversifying generation globally. 

After a high-uncertainty transition is detected, the goal is to probe a small set of immediate alternatives from the same reasoning state, rather than to diversify generation globally. We therefore apply a localized branching procedure that operates directly on the current prefix.

When the uncertainty trigger is activated ($\mathbb{I}_{\text{drift}}(x_t) =1$), GUARD performs short-horizon semantic branching from the fixed prefix $x_{<t}$. A small number of candidate continuations are generated in parallel, each limited to a short horizon $L$. Since all branches share the same prefix, they reuse the pre-computed Key–Value cache of $x_{<t}$, enabling efficient batched generation with minimal latency overhead and only a marginal increase in memory usage. The purpose of branching is to explore distinct local continuations of the same reasoning state, not to approximate an extensive search over solution paths. 

We instantiate three complementary branches. (1) \textbf{Momentum branch}: Generation proceeds from $x_{<t}$ using standard greedy decoding, preserving the model’s current continuation as a reference. (2) \textbf{Inhibitory branch}: The token sequence \texttt{"Wait,"} is prepended before generation, introducing a brief interruption that disrupts immediate continuation patterns. (3) \textbf{Counterfactual branch}: The token sequence \texttt{"Let me reconsider:"} is prepended before generation, encouraging a re-framing of the next reasoning step while retaining the same prefix.

For each branch, we generate a continuation $c^{(i)}$ and evaluate its uncertainty over the generated horizon. This is summarized by the mean token-level entropy
\begin{equation}   
\bar{\mathcal{H}}\!\left(c^{(i)}_t\right)
=
\frac{1}{L}
\sum_{j=0}^{L-1}
\mathcal{H}\!\left(
x_{t+j}^{(i)} \mid x_{<t+j}^{(i)}
\right).
\end{equation}
GUARD selects the continuation with the lowest average entropy,

\begin{equation}
    c^*_t =\arg \min_i \bar{\mathcal{H}} \left( c^{(i)}_t \right),
\end{equation}
and discards the remaining branches. Generation then resumes exclusively from $c^*_t$.

This branch-and-select procedure is deliberately constrained. By confining branching to a short horizon and collapsing back to a single continuation immediately after selection, this procedure probes local alternatives without maintaining parallel trajectories beyond the intervention window.

\subsection{Controlling Late-Stage Reasoning}
\label{subsec:terminate}

Incorrect reasoning trajectories often continue to expand in later stages, whereas correct solutions are typically concise. Once a trajectory has entered a prolonged generation phase, further deliberation is unlikely to reverse an earlier error and instead tends to extend unproductive reasoning. We therefore introduce a lightweight mechanism to control late-stage reasoning and favor timely convergence, at which point the branching mechanism introduced earlier is disabled to prevent further expansion.

We characterize the progression of inference using the remaining capacity ratio,
\begin{equation}
    \rho_t \;=\; 1 - \frac{B_{\text{used}}(t)}{B_{\text{max}}}.
\end{equation}
which measures how far generation has advanced relative to the maximum allowed length. Smaller values of $\rho_t$ correspond to later stages of generation. Termination control is considered only when $\rho_t$ falls below a threshold $\rho_{\text{min}}$, indicating entry into the late stage. 

Within this regime, GUARD monitors the generation stream for hesitation markers that typically precede renewed deliberation. Let $\hat{x}_t$ denote the token predicted by the model at step $t$ and let $\mathcal{T}_{\text{hes}}$ denote a small set of hesitation tokens (e.g., \texttt{"Wait"}). When a hesitation marker is produced in the late stage, GUARD replaces the predicted token with a termination signal,
\begin{equation}
    x_t =
\begin{cases}
\texttt{</think>} 
& \text{if } \hat{x}_t \in \mathcal{T}_{\text{hes}} \;\land\; \rho_t \le \rho_{\min}, \\
\hat{x}_t 
& \text{otherwise}.
\end{cases}
\end{equation}

This design leverages signals already present in the model’s generation behavior to suppress further expansion when additional reasoning is unlikely to be beneficial. Restricting termination control to the late stage preserves flexibility during early reasoning while limiting further expansion once continued deliberation becomes unproductive.

\begin{table*}[!t]
\centering
\caption{\textbf{Performance Comparison Across Multiple Benchmarks.} We report Pass@1 (\%) with the average number of generated tokens shown in parentheses. Standard deviations, when available, are indicated as subscripts. Best and second-best results per benchmark are highlighted with \colorbox{bestBg}{Best} and \colorbox{secondBg}{Second Best}; overall best and second-best averages are marked with \colorbox{avgBestBg}{Best} and \colorbox{avgSecondBg}{Second Best}.
% All results are rounded to one decimal place. When available, standard deviations are indicated as subscripts. For each benchmark, the best and second-best results are highlighted using \colorbox{bestBg}{Best} and \colorbox{secondBg}{Second Best}, respectively. In the Average column, the overall best and second-best methods are distinctly marked with \colorbox{avgBestBg}{Best} and \colorbox{avgSecondBg}{Second Best}.
}
\label{tab:unified_results}

% 优化排版参数
\setlength{\tabcolsep}{3.8pt}
\renewcommand{\arraystretch}{1.25}
\small % 使用小号字体以容纳更多信息

\resizebox{\textwidth}{!}{%
\begin{tabular}{@{}p{1.8cm} *{8}{c} c@{}}
\toprule
% 数据集分类层级（第一层）
\textcolor{categoryText}{\textbf{Method}} & 
\multicolumn{3}{c}{\textcolor{categoryText}{\textsc{\textbf{Competition Reasoning}}}} & 
\multicolumn{2}{c}{\textcolor{categoryText}{\textsc{\textbf{Quantitative}}}} & 
\multicolumn{1}{c}{\textcolor{categoryText}{\textsc{\textbf{Code}}}} & 
\multicolumn{2}{c}{\textcolor{categoryText}{\textsc{\textbf{Domain Knowledge}}}} & 
\multicolumn{1}{c}{\textcolor{categoryText}{\textsc{\textbf{Avg.}}}} \\
\cmidrule(lr){2-4} \cmidrule(lr){5-6} \cmidrule(lr){7-7} \cmidrule(lr){8-9} \cmidrule(lr){10-10}
% 具体数据集名称（第二层）
\textcolor{headerText}{} & 
\textcolor{headerText}{\small \textbf{AIME24}} & 
\textcolor{headerText}{\small \textbf{AIME25}} & 
\textcolor{headerText}{\small \textbf{AMC23}} & 
\textcolor{headerText}{\small \textbf{MATH500}} & 
\textcolor{headerText}{\small \textbf{Minerva}} & 
\textcolor{headerText}{\small \textbf{LiveCode}} & 
\textcolor{headerText}{\small \textbf{Olympiad}} & 
\textcolor{headerText}{\small \textbf{GPQA}} & 
\textcolor{headerText}{\small \textbf{Pass@1}} \\
\midrule
% =========================================================
% Panel A: 1.5B 模型数据
% =========================================================
\multicolumn{10}{c}{\textsc{\textbf{DeepSeek-R1-Distill-Qwen-1.5B}}} \\
\midrule[0.5pt]

BASE & 
\res{20}{0}{8.9k} & \res{13.3}{0}{8.3k} & \res{57}{2.6}{5.8k} & 
\res{78.9}{1.3}{3.8k} & \res{29.5}{0.9}{5.2k} & \res{17.8}{0}{6.9k} & 
\res{39.1}{2.7}{6.3k} & \res{33.8}{0}{7.3k} & \avg{36.2}{6.6k} \\

s1 & 
\res{20}{0}{8.3k} & \res{16.7}{0}{9.1k} & \res{52.5}{0}{6.5k} & 
\res{78.1}{2.5}{5.0k} & \res{32.1}{0.9}{6.1k} & \res{18.4}{0.3}{7.4k} & 
\res{42.1}{1.1}{7.0k} & \res{44.4}{0}{7.9k} & \avg{38.0}{7.2k} \\

CoD & 
\res{16.7}{0}{7.7k} & \res{16.7}{0}{8.5k} & \res{55.8}{2.9}{5.6k} & 
\res{80.2}{0.7}{3.3k} & \res{30.4}{0.5}{4.5k} & \res{19.5}{0.9}{7.2k} & 
\res{41.4}{2.2}{6.0k} & \res{45.5}{0}{6.2k} & \avg{38.3}{6.1k} \\

$\alpha1$ & 
\res{20}{0}{6.8k} & \best{26.7}{0}{6.8k} & \second{70}{0}{4.3k} & 
\res{80.4}{0}{3.5k} & \res{31.2}{0}{4.5k} & \res{21.4}{0.6}{4.8k} & 
\second{44.2}{0.4}{5.0k} & \res{35.9}{0}{4.3k} & \avg{41.2}{5.1k} \\

Reflexion & 
\second{30}{0}{12.8k} & \second{23.3}{0}{12.3k} & \best{72.5}{0}{8.1k} & 
\res{80.2}{1.0}{4.9k} & \second{33.1}{0}{6.9k} & \res{19.3}{1.0}{15.0k} & 
\best{45.5}{0}{9.7k} & \second{46.1}{0}{8.3k} & \avgsecond{43.8}{9.8k} \\

ToT & 
\res{25.5}{13.5}{18.0k} & \res{17.8}{1.9}{17.9k} & \res{58.3}{7.2}{14.6k} & 
\res{74.7}{2.3}{12.1k} & \res{23}{2.3}{12.8k} & \best{22.8}{3.3}{21.9k} & 
\res{38.3}{2.8}{15.1k} & \res{25.8}{14.0}{12.0k} & \avg{35.8}{15.5k} \\

Best-of-N & 
\second{30}{0}{35.0k} & \res{20}{0}{34.4k} & \res{67.5}{0}{23.5k} & 
\best{81.6}{0}{16.1k} & \second{33.1}{0}{22.3k} & \second{21}{0}{7.7k} & 
\res{40.7}{0}{27.4k} & \best{47}{0}{28.7k} & \avg{42.6}{24.4k} \\

Entro-duction & 
\res{16.7}{0}{6.0k} & \res{16.7}{0}{4.5k} & \res{35.8}{10.1}{5.4k} & 
\res{52.2}{0.4}{3.3k} & \res{13.1}{1.7}{4.4k} & \res{18.7}{0.6}{7.1k} & 
\res{20.2}{0.4}{4.2k} & \res{40.4}{0}{5.7k} & \avg{26.7}{5.1k} \\

EAGER & 
\best{33.3}{0}{16.9k} & \second{23.3}{0}{15.5k} & \res{62.5}{0}{11.2k} & 
\res{68.6}{0}{6.6k} & \res{15.4}{0}{8.8k} & \res{17.3}{0.6}{17.5k} & 
\res{30.8}{1.5}{12.8k} & \res{41.8}{0}{16.1k} & \avg{36.6}{13.2k} \\

DTS & 
\res{26.6}{0}{16.8k} & \best{26.7}{0}{17.0k} & \second{70}{0}{10.7k} & 
\res{58.1}{2.2}{6.8k} & \res{22.3}{0.2}{9.4k} & \res{17.8}{0}{16.6k} & 
\res{30.1}{0}{13.6k} & \res{40.6}{1.0}{14.9k} & \avg{36.5}{13.2k} \\
\midrule[0.3pt]
% 我们的方法 (高亮)
\textbf{GUARD} & 
\best{33.3}{0}{9.4k} & \best{26.7}{0}{8.5k} & \best{72.5}{0}{6.5k} & 
\second{81.2}{1.0}{4.8k} & \best{34.6}{0}{6.4k} & \res{20.7}{1.0}{7.7k} & 
\res{43.7}{0}{7.6k} & \best{47}{0}{7.6k} & \avgbest{45.0}{7.3k} \\

\midrule
% =========================================================
% Panel B: 7B 模型
% =========================================================
\multicolumn{10}{c}{\textsc{\textbf{DeepSeek-R1-Distill-Qwen-7B}}} \\
\midrule[0.5pt]

BASE & 
\res{33.3}{0}{8.4k} & \res{26.7}{0}{8.0k} & \res{82.5}{0}{4.7k} & 
\res{87.5}{0.1}{3.4k} & \res{39.7}{0}{4.5k} & \res{43.5}{0}{6.0k} & 
\res{52.6}{0}{6.0k} & \res{44.4}{0}{6.6k} & \avg{51.3}{6.0k} \\

s1 & 
\res{46.7}{0}{8.4k} & \res{26.7}{0}{8.5k} & \res{80}{0}{5.7k} & 
\res{91}{0}{5.6k} & \res{39.7}{0}{5.3k} & \res{44}{0}{6.7k} & 
\res{54.2}{0}{6.7k} & \res{43.9}{0}{7.8k} & \avg{53.3}{6.8k} \\

CoD & 
\res{43.3}{0}{7.5k} & \res{26.7}{0}{7.5k} & \res{85}{0}{3.8k} & 
\res{91}{0}{2.2k} & \res{40.4}{0}{2.2k} & \res{48.3}{0}{6.4k} & 
\res{53.8}{0}{5.0k} & \res{45}{0}{5.3k} & \avg{54.2}{4.9k} \\

$\alpha1$ & 
\res{46.7}{0}{6.8k} & \res{33.3}{0}{6.9k} & \res{82.5}{0}{4.4k} & 
\res{90}{0}{3.9k} & \res{39.7}{0}{4.3k} & \res{48.3}{0}{5.2k} & 
\best{57.5}{0}{5.0k} & \res{47}{0}{4.9k} & \avg{55.6}{5.2k} \\

Reflexion & 
\second{52.2}{3.9}{11.9k} & \best{36.7}{5.8}{12.0k} & \best{90.0}{0}{5.9k} & 
\best{92.6}{0}{5.8k} & \best{42.3}{0}{5.8k} & \res{48.4}{0.1}{11.1k} & 
\best{57.5}{0}{8.3k} & \res{46.1}{2.3}{7.9k} & \avgsecond{58.2}{8.6k} \\

ToT & 
\res{47.8}{3.9}{17.2k} & \second{33.4}{5.8}{17.5k} & \res{78.3}{2.9}{13.4k} & 
\res{87.1}{0.1}{11.5k} & \res{32.5}{1.1}{12.0k} & \best{53.8}{0.4}{19.2k} & 
\res{51.5}{0.7}{14.6k} & \second{51.7}{0.3}{13.8k} & \avg{54.5}{14.9k} \\

Best-of-N & 
\res{36.7}{0}{31.5k} & \res{30}{0}{32.7k} & \res{77.5}{0}{20.0k} & 
\second{91.2}{0}{13.6k} & \res{41.2}{0}{17.9k} & \res{48}{0}{6.7k} & 
\second{55.9}{0}{24.1k} & \res{47}{0}{28.7k} & \avg{53.4}{21.8k} \\

Entro-duction & 
\res{15.6}{2.0}{6.0k} & \res{16.7}{0}{4.5k} & \res{35.8}{10.1}{5.4k} & 
\res{52.2}{0.4}{3.3k} & \res{13.1}{1.7}{4.4k} & \res{18.7}{0.6}{7.1k} & 
\res{20.2}{0.4}{4.2k} & \res{40.4}{0}{5.7k} & \avg{26.6}{5.1k} \\

EAGER & 
\best{60}{0}{10.9k} & \best{36.7}{0}{13.4k} & \best{90}{0}{8.0k} & 
\res{70.6}{0}{5.3k} & \res{25.7}{0}{5.7k} & \res{47.2}{0.3}{17.4k} & 
\res{53}{0}{13.3k} & \res{46}{0}{13.2k} & \avg{53.7}{10.9k} \\

DTS & 
\res{43.3}{0}{13.7k} & \res{26.7}{0}{15.2k} & \best{90}{0}{9.7k} & 
\res{64.8}{0}{4.9k} & \res{33.8}{0}{6.9k} & \res{46.6}{1.0}{12.8k} & 
\res{36.6}{0}{11.1k} & \res{46.4}{0}{11.3k} & \avg{48.5}{10.7k} \\

\midrule[0.3pt]
\textbf{GUARD} & 
\best{60}{0}{8.5k} & \best{36.7}{0}{9.2k} & \second{87.5}{0}{5.8k} & 
\res{90.6}{0}{4.1k} & \second{41.9}{0}{5.3k} & \second{50}{0}{6.6k} & 
\second{55.9}{0}{6.8k} & \best{56.6}{0}{7.4k} & \avgbest{59.8}{6.7k} \\

\midrule
% =========================================================
% Panel C: 32B 模型
% =========================================================
\multicolumn{10}{c}{\textsc{\textbf{Qwen QwQ-32B}}} \\
\midrule[0.5pt]

BASE & 
\res{53.3}{0}{8.7k} & \res{36.7}{0}{8.7k} & \res{77.5}{0}{6.3k} & 
\res{92.4}{0}{4.0k} & \res{46}{0}{5.2k} & \res{73.8}{0}{6.6k} & 
\res{58.8}{0}{6.8k} & \res{56.1}{0}{6.7k} & \avg{61.8}{6.6k} \\

s1 & 
\res{46.7}{0}{8.9k} & \res{43.3}{0}{9.0k} & \res{82.5}{0}{6.6k} & 
\res{91}{0}{4.8k} & \res{48.9}{0}{5.7k} & \res{72}{0}{8.0k} & 
\res{53.4}{0}{7.7k} & \res{43.9}{0}{7.8k} & \avg{60.2}{7.3k} \\

CoD & 
\second{63.3}{0}{7.7k} & \res{46.7}{0}{5.3k} & \res{85}{0}{5.1k} & 
\res{91}{0}{2.8k} & \res{47.4}{0}{3.3k} & \res{76.5}{0}{5.3k} & 
\res{59.9}{0}{5.7k} & \res{56.1}{0}{5.1k} & \avg{65.7}{5.0k} \\

$\alpha1$ & 
\res{53.3}{0}{5.7k} & \res{33.3}{0}{6.3k} & \second{87.5}{0}{4.3k} & 
\res{88.2}{0}{3.2k} & \res{46}{0}{2.9k} & \res{78.25}{0}{6.2k} & 
\res{54.1}{0}{4.5k} & \res{50.5}{0}{3.5k} & \avg{61.4}{4.6k} \\

Reflexion & 
\second{63.3}{0}{12.8k} & \second{50}{0}{14.9k} & \second{87.5}{0}{8.2k} & 
\best{95.2}{0}{4.7k} & \second{50}{0}{7.3k} & \res{79.75}{0}{7.6k} & 
\second{66.2}{0}{10.3k} & \best{59.2}{0}{8.5k} & \avgsecond{68.9}{9.3k} \\

ToT & 
\res{47.8}{3.9}{17.9k} & \res{31.1}{5.1}{18.0k} & \res{85}{0}{15.7k} & 
\res{92}{0}{13.3k} & \res{44.9}{0}{14.3k} & \best{82.4}{6.7}{19.7k} & 
\res{59.3}{0}{16.2k} & \second{58.9}{3.7}{15.0k} & \avg{62.7}{16.3k} \\

Best-of-N & 
\res{55.6}{2.0}{32.4k} & \res{36.7}{0}{32.4k} & \res{78.5}{0}{19.9k} & 
\res{92.4}{0}{13.4k} & \res{47.4}{0}{17.9k} & \res{76.5}{0}{6.8k} & 
\res{59.9}{0}{24.0k} & \res{57.6}{3.5}{27.8k} & \avg{63.1}{21.8k} \\

Entro-duction & 
\res{15.6}{2.0}{6.0k} & \res{16.7}{0}{4.5k} & \res{35.8}{10.1}{5.4k} & 
\res{52.2}{0.4}{3.3k} & \res{13.1}{1.7}{4.4k} & \res{18.7}{0.6}{7.1k} & 
\res{20.2}{0.4}{4.2k} & \res{40.4}{0}{5.7k} & \avg{26.6}{5.1k} \\

EAGER & 
\res{47.8}{1.9}{9.3k} & \res{36.7}{0}{9.8k} & \res{77.5}{0}{6.9k} & 
\res{71.8}{0}{4.6k} & \res{25.7}{0}{6.5k} & \res{76.5}{0}{7.7k} & 
\res{56.0}{3.4}{8.0k} & \res{46}{0}{13.2k} & \avg{54.8}{8.2k} \\

DTS & 
\res{63.3}{0}{14.8k} & \res{46.7}{0}{15.0k} & \best{92.5}{0}{10.4k} & 
\res{84.8}{0}{5.3k} & \res{37.9}{0}{6.9k} & \res{77.75}{0}{12.2k} & 
\res{58.8}{1.9}{12.2k} & \res{53.3}{0}{11.4k} & \avg{64.4}{11.0k} \\
\midrule[0.3pt]
\textbf{GUARD} & 
\best{76.7}{0}{9.2k} & \best{53.3}{0}{9.4k} & \best{92.5}{0}{7.2k} & 
\second{93}{0}{4.9k} & \best{50.4}{0}{6.5k} & \second{80}{0}{6.5k} & 
\best{69.8}{0}{8.9k} & \res{54.5}{0}{7.5k} & \avgbest{71.3}{7.5k} \\

\midrule[1.2pt]
% =========================================================
% Transferability Study: Application on Fine-tuned Model
% =========================================================
\multicolumn{10}{c}{\cellcolor{sectionBg}\textcolor{categoryText}{\textsc{\textbf{Transferability on Math-Specialized Model}}}} \\
\midrule[0.5pt]

\textit{JustRL} & 
\res{40}{0}{7.4k} & \res{24.4}{2.0}{7.2k} & \res{77.5}{0}{5.4k} & 
\res{87.4}{0}{4.0k} & \res{35.7}{0}{5.1k} & \res{17}{0}{7.3k} & 
\res{51}{0}{6.0k} & \res{29.8}{0}{5.0k} & \avg{45.4}{5.9k} \\

\midrule[0.3pt]
\textit{\textbf{+ GUARD}} & 
\best{46.7}{0}{7.8k} & \best{30}{0}{8.0k} & \best{87.5}{0}{5.7k} & 
\res{87.4}{0}{5.0k} & \best{38.6}{0}{6.9k} & \best{32}{0}{7.9k} & 
\best{52.9}{0}{7.3k} & \best{34.8}{0}{7.9k} & \avgbest{51.2}{7.1k} \\

\bottomrule
\end{tabular}%
}
\end{table*}

\section{Experiments}
\subsection{Setup}
\label{sec:experimental_setup}

\textbf{Models and Benchmarks.}
We evaluate our method across model scales using the distilled \textbf{DeepSeek-R1-Distill-Qwen} family (1.5B/7B)~\cite{model-deepseek-r1} and the dense \textbf{QwQ-32B}~\cite{qwen_team_qwq_2025}. 

Experiments are conducted on a diverse benchmark suite spanning four evaluation domains:
(1) \textit{Competition Reasoning}: AMC23~\cite{amc23}, AIME24/25~\cite{aime};
(2) \textit{Formal Quantitative Reasoning}: MATH500~\cite{dataset-math500}, Minerva~\cite{dataset-minerva};
(3) \textit{Coding: LiveCodeBench}~\cite{dataset-livecodebench};
(4) \textit{Domain Knowledge}: OlympiadBench~\cite{dataset-olympiadbench}, GPQA Diamond~\cite{dataset-gpqa}.

% We evaluate efficacy across parameter scales using the distilled \textbf{DeepSeek-R1-Distill} family (1.5B/7B)~\cite{model-deepseek-r1} and the dense \textbf{QwQ-32B}~\cite{qwen_team_qwq_2025}. Our evaluation suite spans four cognitive domains:
% (1) \textbf{\textit{Competition Reasoning}} (AMC23~\cite{amc23}, AIME24/25~\cite{aime}), targeting creative heuristics and multi-step planning;
% (2) \textbf{\textit{Formal Quantitative}} (MATH500~\cite{dataset-math500}, Minerva~\cite{dataset-minerva}), assessing standard academic axioms and symbolic manipulation;
% (3) \textbf{\textit{Coding}} (LiveCodeBench~\cite{dataset-livecodebench}), evaluating procedural logic and algorithmic synthesis; and
% (4) \textbf{\textit{Domain Knowledge}} (OlympiadBench~\cite{dataset-olympiadbench}, GPQA Diamond~\cite{dataset-gpqa}), testing expert-level interdisciplinary synthesis.

\textbf{Evaluation Metrics.}
Following prior work \citep{zhang-etal-2025-alphaone, xu2025dts}, we report \textbf{Pass@1} accuracy and the \textbf{Average Output Length} (in tokens). We present the mean and standard deviation ($\mu \pm \sigma$) across three independent runs.

% \textbf{Implementation Details}
% All experiments use a fixed decoding temperature of 0.0 with top-$p=0.95$, and generation is capped at $T_{\text{max}}=10{,}000$ tokens to reflect realistic deployment budgets. All experiments are conducted on 6 NVIDIA RTX 4090 GPUs. 
% \rev{$q$, $\rho_{\text{min}}$, $\mathcal{T}_{\text{delim}}$, $\mathcal{T}_{\textbf{hes}}$ }

\textbf{Implementation Details.} 
All experiments are conducted on 6 NVIDIA RTX 4090 GPUs using a temperature of 0.0, top-$p=0.95$, and a maximum budget of $B_{\text{max}}=10{,}000$ tokens. For GUARD configurations, we set the entropy quantile $q=0.9$, the late-stage threshold $\rho_{\min}=0.2$, and the branching horizon $L=200$ tokens. The hesitation trigger is set to $\mathcal{T}_{\text{hes}}=\{\texttt{"Wait"}\}$, while $\mathcal{T}_{\text{delim}}$ targets structural boundaries (e.g., \texttt{"\textbackslash n\textbackslash n"}; full list in Appendix~\ref{app:setup}).

\subsection{Main Results}

Table~\ref{tab:unified_results} evaluates GUARD on reasoning-oriented models, comparing it with single-trajectory optimization methods (CoD, s1, $\alpha1$, Reflexion), and parallel search paradigms  (Best-of-N, ToT, Entro-duction, EAGer, DTS).  Detailed configurations are provided in Appendix~\ref{app:baselines}. Across all model scales (1.5B, 7B, and 32B), GUARD consistently achieves the strongest accuracy–length trade-off. In particular, on the 32B model, GUARD attains 71.3\% Pass@1 using only $\sim$7.5k generated tokens, indicating that strong reasoning performance does not require exhaustive parallel sampling or repeated full-chain regeneration.

These gains stem from GUARD’s selective, lightweight intervention. Unlike Reflexion, which relies on external correctness signals and repeated reprocessing that incurs additional inference latency, and parallel search methods, which expand computation and disrupt long-range coherence, GUARD intervenes only at high-risk transitions. By using adaptive, instance-specific uncertainty signals to trigger short-horizon branching, GUARD corrects trajectories efficiently without maintaining parallel paths or relying on fixed thresholds.

% Table~\ref{tab:unified_results} benchmarks GUARD against single-stream optimizations (CoD, s1, $\alpha1$, Reflexion) and parallel search paradigms (Best-of-N, ToT, Entro-duction, EAGer, DTS). Detailed configurations of these baselines are provided in Appendix~\ref{app:baselines}. GUARD establishes a dominant Pareto frontier, consistently achieving the highest average accuracy across 1.5B, 7B, and 32B scales. For instance, on the 32B model, our method attains a leading 71.3\% accuracy using only $\sim$7.5k tokens. This demonstrates that attaining peak capability does not necessitate exhaustive parallel sampling or strictly iterative self-correction.

% The advantage of GUARD stems from architectural efficiency and autonomy. While effective, Reflexion requires oracle signals and incurs overhead by re-processing erroneous chains. Crucially, our metrics exclude prompt costs; Reflexion’s re-ingestion of failures adds significant uncounted latency. Similarly, structured decomposition like Tree-of-Thoughts risks fragmenting the long-context coherence essential for modern Large Reasoning Models. While entropy-aware peers share our focus on uncertainty, they rely on static thresholds that often cause severe degradation. In contrast, GUARD uses adaptive thresholds derived from the model's own historical entropy percentiles to trigger short-horizon semantic branching. Facilitated by KV-cache reuse, this enables efficient in-place rectification without full-context regeneration or brittle hyperparameters.

\begin{table}[t]
\centering

\caption{\textbf{Performance on a General Instruction-Tuned Backbone} Results on Llama-3.1-8B-Instruct compare GUARD with Self-Consistency~\cite{cot-sc2023}, SELF-REFINE~\cite{Iterative-refinement}, and EM-INF~\cite{agarwal2025EM-INF}, highlighting effectiveness beyond reasoning-specialized models. Baseline details are in Appendix~\ref{app:transfer_baselines}.}
\label{tab:llama_results}

% 适应单栏宽度
\resizebox{\columnwidth}{!}{%
\begin{tabular}{l ccccc c}
\toprule
\textbf{Method} & \textbf{Math} & \textbf{AMC} & \textbf{AIME} & \textbf{Minerva} & \textbf{Olymp.} & \textbf{Avg.} \\
\midrule

% --- Baselines ---
Llama-3.1-8B-Instruct & 40.6 & 18.1 & 1.1 & 22.4 & 15.7 & 19.6 \\
Greedy Decoding & 40.6 & 16.9 & 3.3 & 21.0 & 16.0 & 19.6 \\
SELF-REFINE & 41.0 & 19.3 & 1.1 & 22.4 & 15.7 & 19.9 \\
% Self-consistency 在 Olymp 是第二名 (19.4 > 16.6 > 16.4)
Self-consistency & 41.2 & 20.5 & 4.4 & 20.2 & \cellcolor{secondBg}19.4 & 21.1 \\
% Adaptive Temp: Minerva 第一，其他大部分第二
Adaptive Temp & \cellcolor{secondBg}43.6 & \cellcolor{secondBg}25.3 & \cellcolor{secondBg}5.5 & \cellcolor{bestBg}\textbf{24.3} & 16.6 & \cellcolor{avgSecondBg}23.1 \\
EM-INF & 43.0 & 22.9 & 3.3 & 22.8 & 16.4 & 21.7 \\

\midrule

% --- Ours ---
% GUARD: Minerva 第二，其他全部第一
GUARD & \cellcolor{bestBg}\textbf{49.5} & \cellcolor{bestBg}\textbf{32.5} & \cellcolor{bestBg}\textbf{6.7} & \cellcolor{secondBg}23.2 & \cellcolor{bestBg}\textbf{21.7} & \cellcolor{avgBestBg}\textbf{26.7} \\

\bottomrule
\end{tabular}
}
\end{table}

\subsection{Transferability Across Backbone Types}
\label{subsec:transferability}

We further evaluate the generality of GUARD beyond the reasoning-oriented backbones used in our main experiments. Specifically, we consider two complementary settings:
(1) domain-specialized backbones extensively fine-tuned for a specific task, and
(2) general-purpose instruction-tuned backbones without explicit reasoning optimization.
These experiments assess whether GUARD functions as a plug-in inference-time mechanism independent of the backbone’s training paradigm.

\textbf{Math-Specialized Backbones.}
We apply GUARD to JustRL-1.5B~\cite{he2025justrl}, a math-specialized model fine-tuned from \textit{DeepSeek-R1-Distill-Qwen-1.5B}. While such specialization yields strong in-domain performance, it often reduces robustness on non-math tasks. As shown in Table~\ref{tab:unified_results}, GUARD consistently improves mathematical accuracy while also recovering performance on out-of-domain tasks such as coding, indicating that it complements domain-specific fine-tuning through inference-time correction.

\textbf{General Instruction-Tuned Backbones.}
We further evaluate GUARD on Llama-3.1-8B-Instruct and compare it with EM-INF~\cite{agarwal2025EM-INF}, an unsupervised method that reduces entropy globally. In contrast to this global strategy, GUARD intervenes selectively at high-risk transitions. As reported in Table~\ref{tab:llama_results}, GUARD consistently outperforms EM-INF and other baselines (Appendix~\ref{app:transfer_baselines}), demonstrating effectiveness even when the backbone is not optimized for structured reasoning.

\begin{table}[t]
\centering
\caption{\textbf{Ablation Analysis of GUARD.} We report average Pass@1 accuracy across eight benchmarks using DeepSeek-R1-Distill-7B. The $\Delta$ column shows the absolute performance change relative to the full GUARD configuration. }
\label{tab:ablation}
\small
\setlength{\tabcolsep}{8pt} % 垂直排版可以加大列间距，更美观
\begin{tabular}{lcc}
\toprule
\textbf{Configuration} & \textbf{Acc. (\%)} & \textbf{$\Delta$} \\
\midrule
\textbf{Full GUARD}    & \textbf{59.8}      & -      \\
\midrule
\rowcolor{gray!5} \textit{Internal Components} & & \\
\quad w/o Counterfactual & 55.4               & -4.4   \\
\quad w/o Inhibitory     & 57.1              & -2.7   \\
\quad w/o Momentum       & 54.3               & -5.5   \\
\midrule
\rowcolor{gray!5} \textit{Late-stage Reasoning Control} & & \\
\quad w/o Late-stage Control & 53.0               & -6.8   \\
\bottomrule
\end{tabular}
\end{table}

\subsection{Ablation Analysis} \label{subsec:ablation} 

% We report average ablation results on \textbf{DeepSeek-R1-Distill-7B} across eight benchmarks (full details in \textbf{Appendix~\ref{app:ablation_details}}).

% \noindent\textbf{Component Contribution.} Table~\ref{tab:ablation} shows that removing the Momentum, Inhibitory, or Counterfactual branches consistently degrades performance. This indicates that these primitives offer complementary reasoning directions. Their combination is essential to maximize search space coverage and ensure valid rectification paths are found for diverse error types.

% \noindent\textbf{Necessity of Late-Stage Control.} Disabling late-stage intervention causes a sharp performance drop. Without this safeguard, the model generates hesitation markers even when the budget is nearly exhausted, leading to prolonged loops and hard truncation. This confirms that suppressing late-stage hesitation is critical for ensuring the model converges to a valid conclusion.

% \noindent\textbf{Analysis of Hyperparameter Choices.} We further validate our design choices regarding termination thresholds, entropy quantiles, and branching horizons. Detailed hyperparameter sensitivity analyses are provided in \textbf{Appendix~\ref{app:hyperparams}}.

Table~\ref{tab:ablation} reports ablation results on DeepSeek-R1-Distill-Qwen-7B, averaged across eight benchmarks, with full configurations and per-benchmark results provided in Appendix~\ref{app:ablation_details}.

\textbf{Component Contribution.}
All three branching components are necessary for strong performance. Removing any of the Momentum, Inhibitory, or Counterfactual branches consistently degrades accuracy, indicating that effective rectification depends on their complementary roles.

\textbf{Role of Late-Stage Control.}
Late-stage control is critical for preventing performance collapse. Disabling this mechanism leads to prolonged deliberation near the end of generation and a marked drop in performance.

\textbf{Hyperparameter Sensitivity.}
GUARD exhibits stable performance across a wide range of hyperparameter settings. Sensitivity analyses for entropy quantiles, branching horizons, and termination thresholds are reported in Appendix~\ref{app:hyperparams}.

\section*{Conclusion}

% Reasoning performance is often improved by increasing inference-time computation, yet the dynamics of how reasoning fails remain underexplored. We showed that reasoning errors are frequently triggered by a small number of early transitions, propagate through locally coherent reasoning, and are marked by localized entropy spikes, while alternative continuations can still reach correct solutions. Based on these observations, we introduced GUARD, a targeted inference-time framework that intervenes only at high-risk transitions via brief local branching. Our results suggest that understanding where reasoning first deviates is a key complement to scaling-based approaches.

Reasoning performance often improves with increased inference-time computation, yet failure dynamics remain underexplored. We show that errors originate from a few early transitions marked by entropy spikes and propagate through coherent reasoning, motivating GUARD, a targeted inference-time framework that intervenes at high-risk transitions via brief local branching. Our results highlight that identifying where reasoning first deviates complements scaling-based approaches.

\section*{Acknowledgements}
This work was supported by the Joint Key Project of the National Natural Science Foundation of China (U23A20298), the Central Government Fund for Guiding Local Science and Technology Development (202607AD040003), the Key Project of Fundamental Research of Yunnan Province (202401AS070138), the Program of Yunnan Key Laboratory of Intelligent Systems and Computing (202549CE340006), the Yunnan Fundamental Research Project (202501AT070231), the Yunnan University Medical Research Fund (K204209250001), and the Professional Degree Graduate Practice Innovation Project of Yunnan University (ZC-252514097).

% \begin{figure}[t]
%   \includegraphics[width=\columnwidth]{img/fig3_steel_blue_final.pdf}
%   \caption{A figure with a caption that runs for more than one line.
%     Example image is usually available through the \texttt{mwe} package
%     without even mentioning it in the preamble.}
%   \label{fig:experiments}
% \end{figure}

% \begin{figure*}[t]
%   \includegraphics[width=0.48\linewidth]{img/单条熵分析-final_compact_publication_1_v3.pdf} \hfill
%   \includegraphics[width=0.48\linewidth]{img/整体熵分布-distribution_nature_final_fixed.pdf}
%   \caption {A minimal working example to demonstrate how to place
%     two images side-by-side.}
% \end{figure*}

% \begin{figure*}[t]
%   \includegraphics[width=0.48\linewidth]{img/fig1_entropy_dynamics.pdf} \hfill
%   \includegraphics[width=0.48\linewidth]{img/fig2_entropy_distribution.pdf}
%   \caption {A minimal working example to demonstrate how to place
%     two images side-by-side.}
% \end{figure*}

% \clearpage

\section*{Limitations}
Our analysis focuses on trajectory-level failure dynamics under a controlled setup. Segment validity relies on an external oracle and token-level entropy is used as the primary uncertainty signal, which may not capture all forms of reasoning difficulty. Experiments emphasize structured reasoning benchmarks, and how these patterns extend to more open-ended domains or training-time integration remains to be explored.



\bibliography{custom} 

\clearpage

\appendix

% 附录大标题+分隔线
\begin{center}
    \textbf{\Large APPENDIX}
\end{center}
\hrule 
\vspace{1em} 

% 附录目录
\begin{tabular}{@{}l@{\hfill}r@{}} 
    % --- Section A: 算法伪代码 ---
    \textbf{A GUARD Inference Algorithm} & \pageref{app:algorithm} \\ [0.3em]
    
    % --- Section B: 实验设置 ---
    \textbf{B Detailed Experimental Setup} & \pageref{app:setup} \\ 
    \quad B.1 Evaluation Metrics & \pageref{app:metrics} \\ 
    \quad B.2 GUARD Implementation Details & \pageref{app:guard_details} \\  % 新增项
    \quad B.3 Benchmark Details & \pageref{app:benchmarks} \\ 
    \quad B.4 Baseline Descriptions & \pageref{app:baselines} \\ [0.5em] % 顺延编号
    
    % --- Section C: 消融实验 ---
    \textbf{C Additional Ablation Results} & \pageref{app:ablation_details} \\ [0.3em]
    
    % --- Section D: 超参数分析 ---
    \textbf{D Analysis of Hyperparameter Choices} & \pageref{app:hyperparams} \\ [0.3em]
    
    % --- Section E: 案例分析 ---
    \textbf{E Qualitative Analysis of Epistemic Spirals} & \pageref{app:case_study} \\ 
    
    \textbf{F Use of AI Assistants} & \pageref{useai} \\ 

    \textbf{G Artifacts Statements} & \pageref{app:artifacts} \\

\end{tabular}

\vspace{1em} 
\hrule

\section{GUARD Inference Algorithm}
\label{app:algorithm}
Algorithm~\ref{alg:guard} outlines the complete execution workflow of the GUARD framework.

\section{Detailed Experimental Setup}
\label{app:setup}

This appendix provides a comprehensive description of the evaluation metrics, benchmarks, and baseline methods used in our experiments.

\subsection{Evaluation Metrics}
\label{app:metrics}
We employ two primary metrics to assess reasoning performance and computational efficiency:

\paragraph{Pass@1 Accuracy.}
To facilitate consistent evaluation across all models and benchmarks, we explicitly instruct models via the system prompt to enclose their final answer within \texttt{\textbackslash boxed\{\}}. We extract the content inside these tags for verification. For open-ended quantitative tasks (e.g., MATH, AIME), we compare the extracted value against the ground truth using symbolic equivalence checks (e.g., \texttt{sympy}) to account for notational invariance. For multiple-choice tasks (e.g., GPQA), we perform exact string matching on the extracted option key. An output is deemed correct only if the boxed content strictly matches the ground truth label.

\paragraph{Average Output Length.}
To quantify inference efficiency, we measure the total number of tokens generated per query. This includes the entire chain-of-thought reasoning trace and the final answer, but excludes the input prompt tokens. Lower token consumption indicates higher efficiency. For methods involving parallel sampling or tree search, the token count is the sum of tokens generated across all sampled paths or tree branches for a single query.

\subsection{GUARD Implementation Details}
\label{app:guard_details}

In addition to the decoding parameters specified in the main text, we provide the precise definitions of the token sets used for failure detection and intervention triggering.

\paragraph{Hyperparameters.}
The specific thresholds used for the GUARD controller are: entropy quantile sensitivity $q=0.9$, late-stage budget threshold $\rho_{\min}=0.2$, and a short-horizon branching limit of $L=200$ tokens.

\paragraph{Token Definitions.}
The detection mechanism relies on two specific sets of tokens. Note that we represent the newline character as \texttt{\textbackslash n}.

\begin{itemize}
    \item \textbf{Hesitation Set ($\mathcal{T}_{\text{hes}}$).} This set targets explicit linguistic markers of stalling or hesitation generated by the model:
    \begin{equation*}
        \mathcal{T}_{\text{hes}} = \{ \texttt{"Wait"} \}
    \end{equation*}

    \item \textbf{Delimiter Set ($\mathcal{T}_{\text{delim}}$).} This set identifies structural boundaries (e.g., end of paragraphs or logic blocks) where interventions are permitted. It includes standard double-newlines and their combinations with punctuation:
    \begin{equation*}
    \resizebox{.9\hsize}{!}{$
        \mathcal{T}_{\text{delim}} = \left\{
        \begin{aligned}
            & \texttt{"\textbackslash n\textbackslash n"}, \quad \texttt{",\textbackslash n\textbackslash n"}, \quad \texttt{".\textbackslash n\textbackslash n"}, \\
            & \texttt{"]\textbackslash n\textbackslash n"}, \quad \texttt{")\textbackslash n\textbackslash n"}, \quad \texttt{"]),\textbackslash n\textbackslash n"}, \\
            & \texttt{"].\textbackslash n\textbackslash n"}, \quad \texttt{").\textbackslash n\textbackslash n"}, \quad \texttt{".)\textbackslash n\textbackslash n"}
        \end{aligned}
        \right\}
    $}
    \end{equation*}
\end{itemize}

\subsection{Benchmark Details}
\label{app:benchmarks}
Our evaluation suite encompasses four cognitive domains, utilizing datasets specifically chosen for their rigor and ability to differentiate high-capability reasoning models.

\paragraph{Competition Reasoning.}
This category evaluates the model's ability to navigate complex, non-routine problems requiring creative heuristics and multi-step planning.
\begin{itemize}
    \item \textbf{AMC 2023}~\cite{amc23}: A dataset consisting of 40 problems selected from the 2023 AMC 12A and 12B contests. Sponsored by the Mathematical Association of America, these exams target U.S. students in grade 12 and below, featuring challenges across algebra, geometry, number theory, and combinatorics.
    \item \textbf{AIME 2024 \& 2025}~\cite{aime}: A specialized benchmark collection consisting of 60 problems in total—30 from the 2024 American Invitational Mathematics Examination (AIME) and 30 from the 2025 edition. These problems cover core secondary-school mathematics topics but place rigorous demands on both solution accuracy and conceptual depth, serving as a robust test for advanced mathematical reasoning.
\end{itemize}

\paragraph{Formal Quantitative.}
These benchmarks assess the model's command over standard academic axioms and symbolic manipulation.
\begin{itemize}
    \item \textbf{MATH500}~\cite{dataset-math500}: A curated selection of 500 problems extracted from the MATH benchmark. The collection covers a wide range of high-school mathematics domains, including Prealgebra, Algebra, and Number Theory. To ensure comparability with prior work, we utilize the exact problem set originally curated by OpenAI for evaluation.
    \item \textbf{Minerva Math}~\cite{dataset-minerva}: This dataset consists of 272 undergraduate-level STEM problems harvested from MIT’s OpenCourseWare, specifically designed to evaluate multi-step scientific reasoning. The problems span solid-state chemistry, information and entropy, differential equations, and special relativity. Each problem includes a clearly delineated answer—191 verifiable by numeric checks and 81 by symbolic solutions.
\end{itemize}

\paragraph{Coding.}
\begin{itemize}
    \item \textbf{LiveCodeBench}~\cite{dataset-livecodebench}: A contamination-free benchmark for evaluating large language models on code generation. The suite is continuously updated to mitigate data leakage. For this study, we utilize the subset comprising 400 Python programming tasks released between May 2023 and March 2024. Each task is paired with test samples for correctness verification. Beyond basic generation, this benchmark implicitly measures advanced capabilities such as self-repair and edge-case handling.
\end{itemize}

\paragraph{Domain Knowledge.}
This category tests the model's ability to synthesize expert-level knowledge across interdisciplinary fields.
\begin{itemize}
    \item \textbf{OlympiadBench}~\cite{dataset-olympiadbench}: A comprehensive dataset evaluating mathematical and physical reasoning at the Olympiad level. It features a wide difficulty range and expert solution annotations. From the original 8,476 problems, we utilize a specific subset of 675 open-ended, text-only math competition problems in English to focus on pure reasoning without multimodal dependencies.
    \item \textbf{GPQA Diamond}~\cite{dataset-gpqa}: A PhD-level benchmark consisting of high-quality questions spanning physics, chemistry, and biology subdomains. The dataset is notably difficult; domain experts with PhDs in these respective fields achieved only 69.7\% accuracy. We specifically select the highest-quality subset, GPQA Diamond (198 questions), to strictly evaluate the model's capacity for expert-level scientific reasoning and knowledge retrieval.
\end{itemize}

\begin{algorithm}[t]
    \small
    \caption{GUARD Inference Process}
    \label{alg:guard}
    \begin{algorithmic}[1]
        \Require Model $\mathcal{M}$, Prompt $x_{<1}$, Budget $B_{\max}$, Horizon $L$
        \State Initialize $t \leftarrow 0$, sequence $x \leftarrow x_{<1}$, entropy history $\mathbf{H} \leftarrow \emptyset$
        
        \While{$t < B_{\max}$ \textbf{and not} EOS}
            \State Sample candidate $\hat{x}_t$ and compute entropy $h_t$ from $\mathcal{M}(x)$
            \State Update budget ratio $\rho = 1 - t/B_{\max}$
            
            \State \textcolor{gray}{// 1. Late-Stage Control (Sec.~\ref{subsec:terminate})}
            \If{$\rho \le \rho_{\min}$ \textbf{and} $\hat{x}_t \in \mathcal{T}_{\text{hes}}$}
                \State $x \leftarrow x + \texttt{</think>}$ \Comment{Force termination}
                \State \textbf{continue}
            \EndIf
            
            \State \textcolor{gray}{// 2. Failure Detection (Sec.~\ref{subsec:detection})}
            \State Let $x_{last}$ be the last token of $x$
            % 修改点：增加了 \rho > \rho_{\min} 的判断，确保进入 Late-Stage 后不再分支
            \If{$x_{last} \in \mathcal{T}_{\text{delim}}$ \textbf{and} $h_t > \text{Quantile}_q(\mathbf{H})$ \textbf{and} $\rho > \rho_{\min}$}
                \State \textcolor{gray}{// 3. Branch-and-Select (Sec.~\ref{subsec:branch})}
                \State Generate 3 branches $\{c^{(i)}\}_{i=1}^3$ of length $L$:
                \State \quad $\bullet$ \textbf{Momentum}: Greedy from $x$
                \State \quad $\bullet$ \textbf{Inhibitory}: Prepend \texttt{"Wait,"}
                \State \quad $\bullet$ \textbf{Counterfactual}: Prepend \texttt{"Let me re..."} 
                \State Select $c^* = \arg\min_i \bar{\mathcal{H}}(c^{(i)})$ \Comment{Min Entropy}
                \State $x \leftarrow x + c^*$
                \State $t \leftarrow t + L$
            \EndIf

            \State \textcolor{gray}{// Standard Generation}
            \State $x \leftarrow x + \hat{x}_t$
            \State Append $h_t$ to $\mathbf{H}$
            \State $t \leftarrow t + 1$
        \EndWhile
        \State \Return $x$
    \end{algorithmic}
\end{algorithm}

\subsection{Baseline Descriptions}
\label{app:baselines}
We compare GUARD against a wide range of inference-time optimization strategies, categorized into single-stream optimizations and parallel search paradigms.

\subsubsection{Single-Stream Optimizations}
These methods aim to improve reasoning within a single decoding trajectory without maintaining multiple active hypotheses.
\begin{itemize}
    \item \textit{CoD (Chain of Draft)}~\cite{xu2025cod}: A prompting strategy that instructs the model to generate a concise "draft" plan before executing the full reasoning chain. This separates planning from execution to reduce logic errors.
    \item \textit{s1}~\cite{muennighoff-etal-2025-s1}: A budget-forcing method that artificially induces longer deliberation by appending specific wait markers (e.g., "Wait,") to the generation stream. To ensure a fair comparison with other inference-time interventions (following the protocol of $\alpha$1), we apply s1 directly at test-time as a budget-forcing mechanism \textit{without} the supervised fine-tuning (SFT) stage typically associated with its original implementation.
    \item \textit{$\alpha$1}~\cite{zhang-etal-2025-alphaone}: A framework that modulates reasoning duration via a hyperparameter $\alpha$. It treats the insertion of transition tokens as a stochastic process before the $\alpha$ moment, after which it forces deterministic termination of the thought process. We use fixed $\alpha$ values tuned specifically for each benchmark.
    \item \textit{Reflexion}~\cite{shinn2023reflexion}: An iterative self-correction framework where the model critiques and modifies its own output. In our experiments, we employ an oracle-based trigger: reflection is initiated only when the generated answer does not match the ground truth. To strictly align with our evaluation metric of generative token consumption, we report the cumulative sum of output tokens produced across all iteration steps. Crucially, we exclude all prompt tokens (including re-ingested error trajectories and reflection instructions) from this calculation to focus solely on the generative cost.
\end{itemize}

\subsubsection{Parallel Search Paradigms}
These methods leverage computational redundancy to explore a broader solution space.
\begin{itemize}
    \item \textit{Best-of-N (BoN)}~\cite{cot-sc2023}: Adopting the standard self-consistency mechanism, we generate $N=4$ complete independent reasoning paths in parallel for each query. Unlike tree-based methods that evaluate intermediate steps, this approach produces full trajectories before assessment. The final answer is determined via majority voting over the answers extracted from these four parallel candidates.
    \item \textit{ToT (Tree of Thoughts)}~\cite{yao2023tot}: A structured search algorithm that explores the reasoning space by decomposing problems into intermediate steps. We implement ToT using a Depth-First Search (DFS) strategy, where the LLM itself serves as the value function to assign quantitative scores to each intermediate node, guiding the pruning and expansion process. Consistent with the Reflexion baseline, our cost metric accounts solely for the cumulative generated tokens across all visited branches, strictly excluding prompt tokens used for state representation and scoring instructions.
    \item \textit{Entro-duction}~\cite{zhang-etal-2025-entroduction}: A dynamic framework that adjusts reasoning exploration depth by monitoring two uncertainty metrics: the model's output entropy (current step uncertainty) and variance entropy (fluctuation across steps). Based on these signals, the method probabilistically determines whether to deepen the current reasoning path, expand the search space, or terminate exploration. For our implementation, we adhere to the recommended settings with a maximum depth of 20 steps, an exploration rate of 0.25, and a soft-stop buffer of 2 steps.
    \item \textit{EAGER}~\cite{scalena2025eager}: A training-free method that optimizes the efficiency-performance trade-off by dynamically allocating computation based on prompt complexity. Grounded in the assumption that fixed-budget parallel sampling is inefficient for varying problem difficulties, EAGER triggers branching only when detecting high-entropy peaks to concentrate exploration on uncertain steps. While the full framework includes a dataset-level budget reallocation mechanism, we focus on independent per-instance inference. Therefore, we execute only the EAGER-init stage (the preparatory branching phase), using the optimal math configuration: temperature 0.6, entropy threshold 2.2, and a sequence cap of $M=3$.
    \item \textit{DTS} \cite{ xu2025dts}: A framework that constructs a decoding tree by spawning $K$ parallel branches only when the next-token entropy exceeds a threshold $\tau$. A major limitation of the original study is that its efficacy was validated exclusively on the AIME benchmark using fixed hyperparameters, lacking adaptation guidelines for diverse domains. Consequently, we are constrained to applying their fixed threshold ($\tau=2.5$) across all our datasets. Additionally, to ensure a fair comparison regarding computational overhead, we restrict the maximum branching factor to $K=3$.
\end{itemize}

\subsubsection{Baselines for Transferability Analysis}
\label{app:transfer_baselines}

To validate the universality of GUARD across distinct model paradigms (as discussed in Section~\ref{subsec:transferability}), we incorporate two additional baselines representing differing optimization strategies: hyper-specialized RL training and generalist inference-time optimization.

\begin{itemize}
    \item \textit{JustRL}~\cite{he2025justrl}: A minimalist reinforcement learning framework that challenges the necessity of complex multi-stage pipelines. By employing single-stage training with fixed hyperparameters, JustRL achieves state-of-the-art mathematical reasoning performance on 1.5B scale models while using significantly less compute than traditional methods. We utilize the \textbf{JustRL-1.5B} checkpoint (derived from \textit{DeepSeek-R1-Distill-1.5B}) to represent \textit{hyper-specialized models}. Our experiment aims to verify whether GUARD can mitigate the "capability tax"—the degradation of out-of-distribution skills (e.g., coding) often induced by such aggressive domain-specific optimization.

    \item \textit{EM-INF}~\cite{agarwal2025EM-INF}: A specific inference-time variant within the entropy minimization framework that employs logit adjustment to minimize output entropy without parameter updates. Unlike its fine-tuning counterparts, EM-INF requires no labeled data. We select this method as the baseline for generalist models (specifically Llama-3.1-8B-Instruct) because it represents the state-of-the-art for training-free optimization, offering the fairest comparison to our inference-only approach. The reported results for this baseline are directly referenced from the original publication.

    \item \textit{Greedy Decoding}: The standard deterministic generation approach where the sampling temperature is strictly fixed at zero. At each step, the model selects the token with the highest probability, establishing a lower-bound baseline for reasoning stability.
    
    \item \textit{Self-Consistency}~\cite{cot-sc2023}: A parallel sampling strategy designed to marginalize out reasoning errors. Adhering to the evaluation protocol in \cite{agarwal2025EM-INF}, we generate four independent reasoning paths using the prescribed stochastic sampling parameters and derive the final answer via majority voting. This baseline tests whether simple aggregation can outperform targeted steering.
    
    \item \textit{SELF-REFINE}~\cite{Iterative-refinement}: A sequential optimization approach where the model's generated output is fed back into the context to prompt self-correction. We implement this feedback loop for three consecutive iterations following the baseline settings in \cite{agarwal2025EM-INF}, allowing the model to critique and refine its prior outputs.
    
    \item \textit{Adaptive Temperature}: An entropy-aware scaling technique used as a baseline against EM-INF. Instead of using a fixed scalar, this method dynamically reduces the softmax temperature during generation until the output distribution's entropy aligns with the target threshold defined in \cite{agarwal2025EM-INF}. This sharpens the distribution without the direct gradient-based logit updates used in EM-INF.

\end{itemize}

\begin{figure*}[t]
    \centering
    \includegraphics[width=\textwidth]{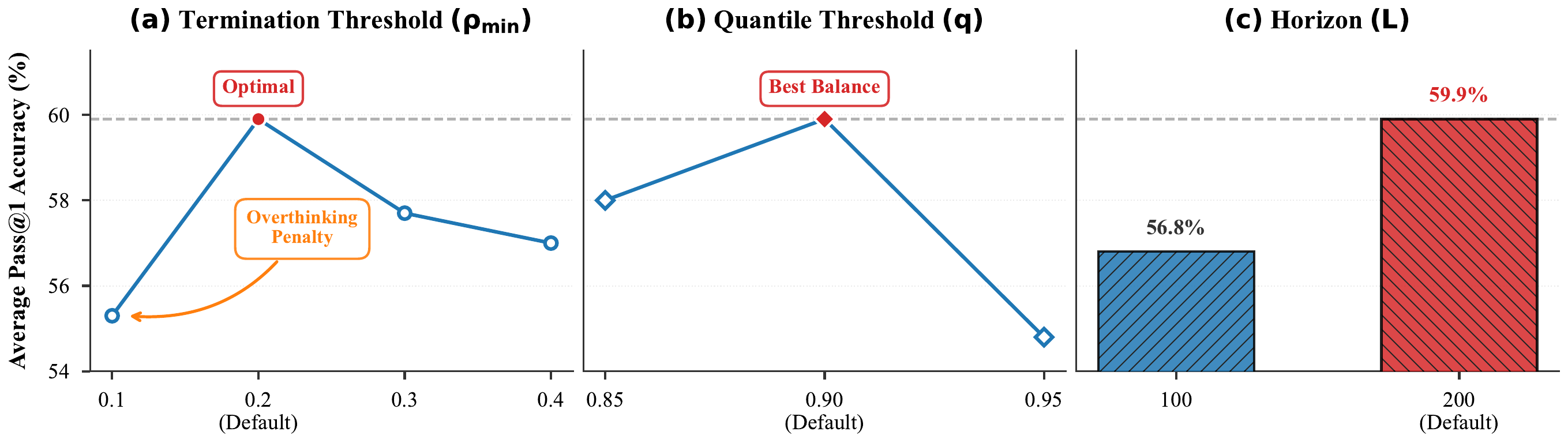}
    \caption{\textbf{Analysis of Hyperparameter Choices.} We analyze how key configuration choices influence model performance.
    \textbf{(a) Termination Threshold $\rho_{\min}$}: Performance peaks at $\rho_{\min}=0.2$. Lower values (0.1) intervene too late, exposing the model to the risk of an "epistemic spiral" in uncontrolled late-stage reasoning.
    \textbf{(b) Quantile Threshold $q$}: $q=0.90$ effectively captures failure onsets without excessive triggering.
    \textbf{(c) Horizon $L$}: A moderate horizon ($L=200$) is chosen as it is sufficient for accurate branch selection, balancing task performance with computational cost.}
    \label{fig:impact_analysis}
\end{figure*}

\begin{table*}[!t]
\centering
\caption{\textbf{Comprehensive Ablation and Sensitivity Analysis.} This table details the performance impact of removing specific branching primitives (Ablation) and varying key hyperparameters (Sensitivity). The \textit{Full Method} (GUARD) adopts the optimal configuration ($L=200, \rho_{\min}=0.2, q=0.90$). Deviating from these settings, such as restricting the horizon, altering termination timing, or changing the quantile threshold, consistently results in suboptimal performance. Results are reported as Pass@1 (\%) with average token usage (k), and all figures are rounded to one decimal place. For each benchmark, the best and second-best results are highlighted using \colorbox{bestBg}{Best} and \colorbox{secondBg}{Second Best}, respectively. In the Average column, the overall best and second-best methods are distinctly marked with \colorbox{avgBestBg}{Best} and \colorbox{avgSecondBg}{Second Best}.}
\label{tab:ablation_full}

% 优化排版参数
\setlength{\tabcolsep}{3.2pt} 
\renewcommand{\arraystretch}{1.2}
\small 

\resizebox{\textwidth}{!}{%
\begin{tabular}{@{}p{3.2cm} *{8}{c} c@{}}
\toprule
% Header
\textcolor{categoryText}{\textbf{Configuration}} & 
\multicolumn{3}{c}{\textcolor{categoryText}{\textsc{\textbf{Competition Reasoning}}}} & 
\multicolumn{2}{c}{\textcolor{categoryText}{\textsc{\textbf{Quantitative}}}} & 
\multicolumn{1}{c}{\textcolor{categoryText}{\textsc{\textbf{Code}}}} & 
\multicolumn{2}{c}{\textcolor{categoryText}{\textsc{\textbf{Domain Knowledge}}}} & 
\multicolumn{1}{c}{\textcolor{categoryText}{\textsc{\textbf{Avg.}}}} \\
\cmidrule(lr){2-4} \cmidrule(lr){5-6} \cmidrule(lr){7-7} \cmidrule(lr){8-9} \cmidrule(lr){10-10}
\textcolor{headerText}{} & 
\textcolor{headerText}{\small \textbf{AIME24}} & 
\textcolor{headerText}{\small \textbf{AIME25}} & 
\textcolor{headerText}{\small \textbf{AMC23}} & 
\textcolor{headerText}{\small \textbf{MATH500}} & 
\textcolor{headerText}{\small \textbf{Minerva}} & 
\textcolor{headerText}{\small \textbf{LiveCode}} & 
\textcolor{headerText}{\small \textbf{Olympiad}} & 
\textcolor{headerText}{\small \textbf{GPQA}} & 
\textcolor{headerText}{\small \textbf{Pass@1}} \\
\midrule

% =========================================================
% Ablation: Branching Primitives
% =========================================================
\multicolumn{10}{l}{\textit{\textbf{Ablation: Branching Primitives}}} \\
\midrule[0.3pt]

w/o Momentum & 
\res{46.7}{0}{7.5k} & \res{30.0}{0}{8.2k} & \res{77.5}{0}{5.3k} & 
\res{86.4}{0}{4.2k} & \res{41.2}{0}{4.9k} & \res{47.2}{0}{5.6k} & 
\res{50.6}{0}{6.6k} & \res{55.0}{0}{5.2k} & \avg{54.3}{5.9k} \\

w/o Inhibitory & 
\res{46.7}{0}{7.3k} & \best{36.7}{0}{7.9k} & \res{80.0}{0}{5.1k} & 
\res{86.4}{0}{3.9k} & \second{42.6}{0}{4.8k} & \res{48.3}{0}{5.7k} & 
\res{51.7}{0}{6.2k} & \best{64.7}{0}{5.2k} & \avg{57.1}{5.8k} \\

w/o Counterfactual & 
\res{46.7}{0}{7.4k} & \res{30.0}{0}{7.9k} & \res{80.0}{0}{5.2k} & 
\res{87.2}{0}{4.0k} & \res{40.1}{0}{4.8k} & \res{48.6}{0}{5.6k} & 
\res{51.3}{0}{6.2k} & \second{59.1}{0}{6.4k} & \avg{55.4}{5.9k} \\

\midrule
% =========================================================
% Sensitivity: Horizon
% =========================================================
\multicolumn{10}{l}{\textit{\textbf{Sensitivity: Branching Horizon ($L$)}}} \\
\midrule[0.3pt]

Horizon $L=100$ & 
\res{50.0}{0}{8.0k} & \best{36.7}{0}{8.2k} & \res{80.0}{0}{4.8k} & 
\res{89.2}{0}{3.1k} & \res{40.4}{0}{4.1k} & \res{48.3}{0}{5.6k} & 
\res{53.5}{0}{5.9k} & \res{56.6}{0}{6.2k} & \avg{56.8}{5.7k} \\

\midrule
% =========================================================
% Sensitivity: Termination Threshold (Rho_min)
% =========================================================
\multicolumn{10}{l}{\textit{\textbf{Sensitivity: Termination Threshold ($\rho_{\min}$)}}} \\
\midrule[0.3pt]

Threshold $\rho_{\min}=0.1$ & 
\res{46.7}{0}{8.5k} & \res{26.7}{0}{9.9k} & \res{82.5}{0}{6.8k} & 
\res{88.2}{0}{5.1k} & \best{45.2}{0}{5.7k} & \res{46.1}{0}{7.8k} & 
\res{53.8}{0}{7.2k} & \res{53.0}{0}{8.2k} & \avg{55.3}{7.4k} \\

Threshold $\rho_{\min}=0.3$ & 
\second{53.3}{0}{8.3k} & \res{30.0}{0}{9.2k} & \best{90.0}{0}{5.5k} & 
\res{88.8}{0}{4.1k} & \res{41.5}{0}{5.1k} & \res{48.3}{0}{6.4k} & 
\res{54.1}{0}{6.8k} & \res{55.6}{0}{7.1k} & \avg{57.7}{6.6k} \\

Threshold $\rho_{\min}=0.4$ & 
\res{43.3}{0}{8.7k} & \res{30.0}{0}{8.9k} & \best{90.0}{0}{4.6k} & 
\second{90.4}{0}{3.8k} & \second{42.6}{0}{4.7k} & \res{48.3}{0}{6.2k} & 
\second{55.9}{0}{5.5k} & \res{55.6}{0}{7.1k} & \avg{57.0}{6.2k} \\

w/o Late-Stage Control & 
\res{36.7}{0}{9.1k} & \res{30.0}{0}{9.9k} & \res{80.0}{0}{7.1k} & 
\res{86.4}{0}{5.6k} & \second{42.6}{0}{6.8k} & \res{45.5}{0}{8.1k} & 
\res{51.7}{0}{7.8k} & \res{51.0}{0}{8.9k} & \avg{53.0}{7.9k} \\

\midrule
% =========================================================
% Sensitivity: Quantile Threshold (q)
% =========================================================
\multicolumn{10}{l}{\textit{\textbf{Sensitivity: Quantile Threshold ($q$)}}} \\
\midrule[0.3pt]

Quantile $q=0.85$ & 
\second{53.3}{0}{9.5k} & \res{33.3}{0}{9.4k} & \res{85.0}{0}{6.1k} & 
\best{90.6}{0}{7.0k} & \res{40.8}{0}{6.1k} & \best{50.0}{0}{8.0k} & 
\best{56.6}{0}{7.0k} & \res{54.0}{0}{8.5k} & \avgsecond{58.0}{7.7k} \\

Quantile $q=0.95$ & 
\res{36.7}{0}{7.9k} & \res{30.0}{0}{7.5k} & \second{87.5}{0}{5.3k} & 
\res{88.0}{0}{4.0k} & \res{41.9}{0}{5.1k} & \second{49.2}{0}{5.3k} & 
\res{55.6}{0}{6.2k} & \res{49.5}{0}{6.7k} & \avg{54.8}{6.0k} \\

\midrule[1.0pt]
% =========================================================
% Full Method (Reference)
% =========================================================
\textbf{GUARD (Full Method)} & 
\best{60}{0}{8.5k} & \best{36.7}{0}{9.2k} & \second{87.5}{0}{5.8k} & 
\best{90.6}{0}{4.1k} & \res{41.9}{0}{5.3k} & \best{50}{0}{6.6k} & 
\second{55.9}{0}{6.8k} & \res{56.6}{0}{7.4k} & \avgbest{59.8}{6.7k} \\

\bottomrule
\end{tabular}%
}
\end{table*}

\section{Additional Ablation Results}
\label{app:ablation_details}

In this section, we provide the fine-grained numerical breakdown of the ablation study summarized in Section~\ref{subsec:ablation}. Table~\ref{tab:ablation_full} details the performance of GUARD on the \textbf{DeepSeek-R1-Distill-Qwen-7B} model across all eight benchmarks when individual branching primitives are removed.

\paragraph{Impact of Branching Primitives.}
Consistent with the aggregated results in the main text, we observe that removing any single branch type—\textit{Momentum}, \textit{Inhibitory}, or \textit{Counterfactual}—results in performance degradation across the majority of domains. This reinforces our hypothesis that these strategies probe distinct reasoning subspaces:
\begin{itemize}
    \item The \textbf{Momentum branch} leverages the model's intrinsic generation inertia to preserve valid partial reasoning.
    \item The \textbf{Inhibitory branch} (induced by "Wait,") effectively disrupts premature convergence and "system 1" thinking patterns.
    \item The \textbf{Counterfactual branch} (induced by "Let me reconsider:") explicitly encourages the model to explore alternative logical paths from the same context.
\end{itemize}
The results demonstrate that the synergy of these three primitives is essential for maximizing the coverage of the search space, ensuring that valid rectification paths are discovered across diverse failure modes—from symbolic manipulation errors in MATH/Minerva to logic flaws in LiveCodeBench.

\section{Analysis of Hyperparameter Choices}
\label{app:hyperparams}

We characterize the behavior of GUARD under varying configurations to validate our design choices regarding the termination threshold ($\rho_{\min}$), quantile threshold ($q$), and horizon length ($L$). As shown in Figure~\ref{fig:impact_analysis}, our default configuration represents a balanced operating point that maximizes accuracy while maintaining inference efficiency. Detailed numerical results for all hyperparameter configurations are provided in Table~\ref{tab:ablation_full}.

\paragraph{Termination Threshold ($\rho_{\min}$).}
Figure~\ref{fig:impact_analysis}(a) validates our choice of $\rho_{\min}=0.2$.
The model achieves peak performance ($59.9\%$) when control is activated upon entering the final $20\%$ of the budget. In contrast, delaying intervention ($\rho_{\min}=0.1$) causes a sharp performance drop to $55.3\%$. This confirms that unconstrained deliberation in the final stages incurs an "epistemic spiral," necessitating accelerated convergence. Conversely, activating control too early ($\rho_{\min} \ge 0.3$) also reduces accuracy ($57.7\%$) by limiting the reasoning depth required for complex problems.

\paragraph{Quantile Threshold ($q$) \& Horizon ($L$).}
Figures~\ref{fig:impact_analysis}(b) and (c) justify our selection of branching parameters. A quantile of $q=0.90$ provides the most effective signal-to-noise ratio for failure detection. Regarding the horizon, we adopt $L=200$. While significantly longer horizons might offer theoretical benefits, we observe that $L=200$ is sufficient for reliable entropy estimation. We deliberately chose not to increase $L$ further to preserve the lightweight nature of our inference-time intervention.

\section{Qualitative Analysis of Epistemic Spirals}
\label{app:case_study}

We present a qualitative comparison across three model scales (1.5B, 7B, 32B) and domains (Geometry, Physics, Number Theory) to demonstrate GUARD's versatility. We observe that "Epistemic Spirals" manifest differently depending on model capability. As shown below, GUARD identifies these failures via entropy spikes and intervenes with context-aware branching to restore convergence.

\paragraph{1. Overcoming Arithmetic Hesitation (Figure~\ref{fig:case_study_aime}).}
In high-precision geometry (AIME 2024), smaller models like \textbf{DeepSeek-R1-Distill-Qwen-1.5B} often falter when facing complex arithmetic. As shown in Figure~\ref{fig:case_study_aime}, the model derives the correct equations but enters a loop of self-doubt due to large coefficients ($>10^7$). GUARD detects this hesitation and injects a Counterfactual Branch, enforcing the execution of the calculation to reveal the integer solution that the base model initially abandoned.

\paragraph{2. Resolving Intuition Conflicts (Figure~\ref{fig:case_study_minerva}).}
In physics reasoning (Minerva), even capable models like \textbf{DeepSeek-R1-Distill-Qwen-7B} struggle when correct results contradict training priors. Figure~\ref{fig:case_study_minerva} illustrates a case where the model doubts a valid but counter-intuitive result (a macroscopic atomic wavelength), triggering unnecessary error checking. GUARD intervenes with a Scaling Law Verification, guiding the model to validate the result via first-principles estimation rather than rejecting the correct path.

\paragraph{3. Shifting form Brute-force to Structure (Figure~\ref{fig:case_study_olympiad}).}
In number theory (OlympiadBench), larger models like \textbf{QwQ-32B} may attempt to solve structural problems through inefficient enumeration. As depicted in Figure~\ref{fig:case_study_olympiad}, the base model wastes tokens searching for non-existent counter-examples. GUARD detects the lack of logical progression and injects a Counterfactual Branch, steering the model away from aimless guessing toward a rigorous proof based on modular arithmetic.

\section{Use of AI Assistants}
\label{useai}
We utilized AI assistants to help with language editing and writing refinement. All technical content, experimental results, and scientific claims were verified by the authors.

\section{Artifacts Statements}
\label{app:artifacts}

\subsection{Model Artifacts}
We utilize the following models in our work, complying with all respective license terms:

\begin{itemize}
    \item DeepSeek-R1-Distill-Qwen-1.5B and DeepSeek-R1-Distill-Qwen-7B: Both models are released under the MIT License, which permits commercial use, modification, and redistribution. These models are distilled from the Qwen-2.5 series (Apache 2.0 License).
    
    \item Qwen QwQ-32B: This model is released under the Apache License 2.0, allowing both research and commercial usage.
    
    \item Llama-3.1-8B-Instruct: Used for our generalist model transferability analysis, this model is released under the Llama 3.1 Community License. We comply with the usage policy and acceptable use guidelines provided by Meta.
\end{itemize}

\subsection{Data Artifacts}
We employ publicly available benchmarks for evaluation, including MATH-500, LiveCodeBench, OlympiadBench, GPQA, AIME, AMC 23, and Minerva. These datasets are widely accessible in the open-source community. We utilize them strictly for non-commercial research purposes and ensure that our usage complies with the respective licenses and terms of use.

\subsection{External Services}
For automated evaluation, we utilize the Gemini 3 Pro API. We comply with the Google AI Studio Terms of Service regarding data handling and API usage limits.

\begin{figure*}[t]
\centering
\begin{tcolorbox}[
    enhanced,
    title={\textsc{Case Study: DeepSeek-R1-Distill-Qwen-1.5B on AIME}},
    colframe=mainFrame, colback=white, coltitle=white,
    fonttitle=\bfseries\large,
    boxrule=1.0pt,
    drop shadow,
    left=10pt, right=10pt, top=10pt, bottom=10pt
]

    % --- 问题描述 ---
    \textbf{Question:} 
    Let $ABC$ be a triangle inscribed in circle $\omega$. Tangents at $B$ and $C$ intersect at $D$, and $AD$ intersects $\omega$ at $P$. If $AB=5$, $BC=9$, $AC=10$, find $m+n$ where $AP = m/n$.
    \vspace{0.3cm}

    % --- 错误案例 (Base Model) ---
    \begin{tcolorbox}[
        enhanced,
        title={\textbf{\textcolor{white}{Base Model (Failure): Calculation Anxiety \& Spiral}}},
        colframe=failFrame, colback=failBack,
        coltitle=white,
        attach boxed title to top left={yshift=-2mm, xshift=2mm},
        boxed title style={boxrule=0pt, colframe=failFrame},
        boxrule=0.5pt,
        sharp corners=south
    ]
    \small
   \textbf{Reasoning Trace:}\\
    \textit{[...Step 1: Coordinates setup $A(1/3, 4\sqrt{14}/3)$, $D(9/2, -27\sqrt{14}/11)$...]}\\
    \textit{[...Step 2: Derives quadratic for intersection...]}
    
    The equation is $218750 (33 - 168t)^2 / 5544^2 + \dots = 765/112$.
    \txtspiral{The numbers are getting huge. $5544^2$ is around 30 million. This doesn't seem right.}
    \txtspiral{Let me double check the slope of $OB$. Did I miss a negative sign? Let me recalculate...}
    \txtspiral{Slope seems correct. Maybe I should use Stewart's Theorem instead? But I need length $AD$...}
    \txtspiral{Let's go back to coordinates. Is it possible that $y_A$ is negative? No...}
    \txtspiral{Let me try to factor 218750... No, this is taking too long.}
    \textit{[...Model abandons the correct coordinate path and loops between verifying arithmetic and trying new methods until cutoff...]}
    
    \textbf{Final Answer:} \textcolor{failFrame}{Timeout / No Solution}
    \end{tcolorbox}
    \vspace{0.2cm}

    % --- 正确案例 (GUARD) ---
    \begin{tcolorbox}[
        enhanced,
        title={\textbf{\textcolor{white}{GUARD (Success): Confidence Injection}}},
        colframe=succFrame, colback=succBack,
        coltitle=white,
        attach boxed title to top left={yshift=-2mm, xshift=2mm},
        boxed title style={boxrule=0pt, colframe=succFrame},
        boxrule=0.5pt,
        sharp corners=south
    ]
    \small
    \textbf{Reasoning Trace:}\\
    \textit{[...Coordinates setup identical to Base Model...]}\\
    We arrive at the quadratic: $1521x^2 - 4314x + 1269 = 0$.
    The coefficients are large. I need to solve for $x$.
    
    % 这里体现：\n\n 后紧跟的高熵 Token 触发了检测
    \texttt{\textbackslash n\textbackslash n} \txtspiral{This...} \hfill \textbf{\textit{\textcolor{gray}{[$\uparrow$ Entropy Spike Detected after delimiter]}}}
    
    \vspace{0.1cm}
    \textbf{\textcolor{succFrame}{$\Rightarrow$ GUARD Triggered: Branch 3 (Counterfactual) Selected}}
    
    \txtcorrect{Let me reconsider: Instead of doubting the size, let's calculate the discriminant $D$ carefully to see if it's a perfect square.}
    
    $D = b^2 - 4ac = 4314^2 - 4(1521)(1269)$.
    $4314^2 = 18,610,596$.
    $4(1521)(1269) = 6084 \times 1269 = 7,718,596$.
    $D = 18,610,596 - 7,718,596 = 10,892,000$.
    
    Wait, $10,892,000 = 108920 \times 100$? No, $1089 \times 10000$.
    $\sqrt{D} = \sqrt{1089 \times 10000} = 33 \times 100 = 3300$.
    \textbf{It is a perfect square!}
    
    Roots: $x = \frac{4314 \pm 3300}{3042}$.
    Solution $P$: $x = \frac{7614}{3042} = \frac{423}{169}$.
    Calculating distance $AP$:
    $AP = \sqrt{(423/169 - 1/3)^2 + \dots} = \frac{100}{13}$.
    
    Thus, $m=100, n=13$. $m+n=113$.
    
    \textbf{Final Answer:} \txtcorrect{113}
    \end{tcolorbox}

    % --- 分析 ---
    \vspace{0.2cm}
    \noindent\textbf{Analysis:} This case illustrates an "Arithmetic Spiral." The Base model correctly derives the coordinate equations but panics due to the magnitude of the coefficients ($>10^7$), interpreting the complexity as a sign of error (red text). It abandons the correct path to search for non-existent shortcuts. GUARD detects this hesitation and intervenes with a Counterfactual Branch that encourages verification rather than abandonment. By pushing through the discriminant calculation, the model discovers the numbers simplify beautifully (green text), resolving the problem.

\end{tcolorbox}
\caption{\textbf{Qualitative Comparison on AIME.} The Base model (Top) retreats from a correct but computationally heavy solution at the structural boundary \texttt{\textbackslash n\textbackslash n}. GUARD (Bottom) detects the uncertainty spike immediately following the delimiter and steers the model to complete the calculation.}
\label{fig:case_study_aime}
\end{figure*}

\begin{figure*}[t]
\centering
\begin{tcolorbox}[
    enhanced,
    title={\textsc{Case Study: DeepSeek-R1-Distill-Qwen-7B on Minerva}},
    colframe=mainFrame, colback=white, coltitle=white,
    fonttitle=\bfseries\large,
    boxrule=1.0pt,
    drop shadow,
    left=10pt, right=10pt, top=10pt, bottom=10pt
]

    % --- 问题描述 ---
    \textbf{Question:} 
    Use the Bohr model of the atom to compute the wavelength of the transition from the $n=100$ to $n=99$ levels, in centimeters. [Useful relation: the wavelength of $L\alpha$ ($n=2$ to $n=1$ transition) is $1216 \text{\AA}$].
    \vspace{0.3cm}

    % --- 错误案例 (Base Model) ---
    \begin{tcolorbox}[
        enhanced,
        title={\textbf{\textcolor{white}{Base Model (Failure): Intuition Trap \& Unit Spiral}}},
        colframe=failFrame, colback=failBack,
        coltitle=white,
        attach boxed title to top left={yshift=-2mm, xshift=2mm},
        boxed title style={boxrule=0pt, colframe=failFrame},
        boxrule=0.5pt,
        sharp corners=south
    ]
    \small
    \textbf{Reasoning Trace:}\\
    First, I need to find the Rydberg constant $R$. Using the $L\alpha$ transition ($n=2 \to 1$):
    $1/\lambda = R(1/1^2 - 1/2^2) = 0.75R$.
    Given $\lambda = 1216 \text{\AA} = 1.216 \times 10^{-5} \text{ cm}$, so $R = 4 / (3 \times 1.216 \times 10^{-5}) \approx 1.0965 \times 10^5 \text{ cm}^{-1}$.
    
    Now for the transition $n=100 \to 99$:
    $1/\lambda' = R(1/99^2 - 1/100^2)$.
    $1/99^2 - 1/100^2 = \frac{10000 - 9801}{98010000} = \frac{199}{9.8 \times 10^7} \approx 2.03 \times 10^{-6}$.
    So $1/\lambda' = 1.0965 \times 10^5 \times 2.03 \times 10^{-6} \approx 0.222 \text{ cm}^{-1}$.
    Therefore, $\lambda' = 1 / 0.222 \approx 4.49 \text{ cm}$.
    
    \texttt{\textbackslash n\textbackslash n}
    \txtspiral{4.49 cm? That seems impossibly long.}
    \txtspiral{The L-alpha line is UV radiation ($10^{-5}$ cm). How can an atomic transition be in the microwave range (centimeters)? Usually, atomic spectra are visible or UV.}
    
    \texttt{\textbackslash n\textbackslash n}
    \txtspiral{I must have messed up the units of the Rydberg constant. Let me check.}
    \txtspiral{Maybe $1216 \text{\AA}$ should be converted to meters first? $1.216 \times 10^{-7}$ m.}
    \txtspiral{Re-calculating R in $m^{-1}$: $R \approx 1.097 \times 10^7 m^{-1}$.}
    \txtspiral{Plugging back: $1/\lambda' \approx 22.2 m^{-1} \implies \lambda' \approx 0.045$ m. }
    \txtspiral{0.045 m is still 4.5 cm. This doesn't make sense. Atoms are tiny, how can they emit centimeter-long waves?}
    
    \texttt{\textbackslash n\textbackslash n}
    \txtspiral{Let me try using the energy formula $E = -13.6 \text{ eV} / n^2$.}
    \txtspiral{$\Delta E = 13.6(1/99^2 - 1/100^2) \approx 2.7 \times 10^{-5}$ eV.}
    \txtspiral{Convert eV to wavelength... $12400 / (2.7 \times 10^{-5}) \approx 4.5 \times 10^8 \text{\AA}$.}
    \txtspiral{That is HUGE. I must be wrong. Is it possible the question implies $Z$ is not 1?}
    \textit{[...Model continues to loop through different constants (Planck, speed of light) trying to "fix" the large result, eventually hallucinating a scaling factor...]}
    
    \textbf{Final Answer:} \textcolor{failFrame}{Invalid / Timeout}
    \end{tcolorbox}
    \vspace{0.2cm}

    % --- 正确案例 (GUARD) ---
    \begin{tcolorbox}[
        enhanced,
        title={\textbf{\textcolor{white}{GUARD (Success): Scaling Law Verification}}},
        colframe=succFrame, colback=succBack,
        coltitle=white,
        attach boxed title to top left={yshift=-2mm, xshift=2mm},
        boxed title style={boxrule=0pt, colframe=succFrame},
        boxrule=0.5pt,
        sharp corners=south
    ]
    \small
    \textbf{Reasoning Trace:}\\
    \textit{[...Initial calculation matches Base Model...]}
    Using $R \approx 1.097 \times 10^5 \text{ cm}^{-1}$, we derived $1/\lambda' \approx 0.222 \text{ cm}^{-1}$.
    This yields $\lambda' \approx 4.49 \text{ cm}$.
    
    \texttt{\textbackslash n\textbackslash n} \txtspiral{this result seems...} \hfill \textbf{\textit{\textcolor{gray}{[$\uparrow$ Entropy Spike Detected]}}}
    
    \vspace{0.1cm}
    \textbf{\textcolor{succFrame}{$\Rightarrow$ GUARD Triggered: Branch 2 (Inhibitory) "Wait,"}}
    
    \txtcorrect{Wait, let's not discard the result yet. Let's perform an order-of-magnitude estimation to see if this is physically plausible for high-$n$ states (Rydberg states).}
    
    1. Energy Scaling: Energy levels scale as $1/n^2$.
       The transition $n=2 \to 1$ involves a large gap: $(1 - 1/4) = 0.75$.
       The transition $n=100 \to 99$ involves a tiny gap: $\approx 2/n^3 \approx 2/1,000,000 = 2 \times 10^{-6}$.
    
    2. Ratio Comparison:
       Ratio of Energy gaps $\approx 0.75 / (2 \times 10^{-6}) \approx 3.75 \times 10^5$.
       Since Wavelength $\lambda \propto 1/\Delta E$, the new wavelength should be roughly $3.75 \times 10^5$ times longer than $L\alpha$.
    
    3. Verification:
       $\lambda_{L\alpha} = 1216 \text{\AA} \approx 1.2 \times 10^{-5} \text{ cm}$.
       Expected $\lambda' \approx (1.2 \times 10^{-5} \text{ cm}) \times (3.75 \times 10^5) \approx 4.5 \text{ cm}$.
    
    This matches our calculated result of $4.49 \text{ cm}$. High-$n$ transitions indeed produce low-energy photons (microwaves/radio waves). The calculation is correct.
    
    \textbf{Final Answer:} \txtcorrect{4.49 cm}
    \end{tcolorbox}

    % --- 分析 ---
    \vspace{0.2cm}
   \noindent\textbf{Analysis:} The Base model enters an epistemic spiral triggered by the dissonance between its correct calculation ($4.49$ cm) and a faulty intuition. Driven by this misconception, the model misleads itself, engaging in a desperate search to prove its correct answer wrong. It repeatedly cycles through unit conversions and constant checks not to solve the problem, but to find a non-existent "bug" that would justify its intuition. GUARD interrupts this self-defeating loop by enforcing a physics-based scaling check, forcing the model to accept the counter-intuitive truth rather than spiraling into hallucination.

\end{tcolorbox}
\caption{\textbf{Qualitative Comparison on Minerva.} The Base model (Top) correctly computes the value but actively rejects the result due to its counter-intuitive magnitude, entering a spiral to rationalize its error. GUARD (Bottom) breaks this self-defeating loop by injecting a scaling-law verification, confirming the physical validity of the result.}
\label{fig:case_study_minerva}
\end{figure*}

\begin{figure*}[t]
\centering
\begin{tcolorbox}[
    enhanced,
    title={\textsc{Case Study: QwQ-32B on OlympiadBench}},
    colframe=mainFrame, colback=white, coltitle=white,
    fonttitle=\bfseries\large,
    boxrule=1.0pt,
    drop shadow,
    left=10pt, right=10pt, top=10pt, bottom=10pt
]

    % --- 问题描述 ---
    \textbf{Question:} 
    Define sequence $x_{1}=a, x_{n+1}=2 x_{n}+1$. Let $y_{n}=2^{x_{n}}-1$. Determine the largest $k$ such that $y_{1}, \ldots, y_{k}$ are all prime for some positive integer $a$.
    \vspace{0.3cm}

    % --- 错误案例 (Base Model) ---
    \begin{tcolorbox}[
        enhanced,
        title={\textbf{\textcolor{white}{Base Model (Failure): The Empirical Spiral}}},
        colframe=failFrame, colback=failBack,
        coltitle=white,
        attach boxed title to top left={yshift=-2mm, xshift=2mm},
        boxed title style={boxrule=0pt, colframe=failFrame},
        boxrule=0.5pt,
        sharp corners=south
    ]
    \small
    \textbf{Reasoning Trace:}\\
    Let's test small values of $a$.
    If $a=2$: $x=\{2,5,11\}$, $y=\{3,31,2047\}$. $2047=23 \times 89$ (composite). So $k=2$.
    If $a=3$: $x=\{3,7,15\}$, $y=\{7,127,32767\}$. $32767$ is composite. So $k=2$.
    
    \texttt{\textbackslash n\textbackslash n}
    \txtspiral{Is it possible to find $k=3$ with a larger $a$? We need $x_1, x_2, x_3$ to be Mersenne exponents.}
    \txtspiral{Let's check $a=5$... $x_2=11$, $y_2=2047$ (composite). Fail.}
    \txtspiral{Let's check $a=7$... $x_2=15$ (composite). Fail.}
    \txtspiral{Let's check $a=13$... $x_2=27$ (composite). Fail.}
    
    \texttt{\textbackslash n\textbackslash n}
    \txtspiral{Maybe $a=89$? $y_1$ is prime. $x_2=179$. Is $2^{179}-1$ prime? That's too big to check mentally.}
    \txtspiral{Let me try finding patterns modulo 3. If $p_1 \equiv 1$, then $p_2 \equiv 0$ (bad). If $p_1 \equiv 2$, then $p_2 \equiv 2, p_3 \equiv 2$. This doesn't rule out $k=3$.}
    \txtspiral{What about modulo 7? Modulo 5? I can't find a contradiction.}
    \txtspiral{Maybe $a=127$? Or $a=19937$? There must be some $a$ that works...}
    \textit{[...Model spirals into an infinite search for a counter-example without a theoretical proof strategy...]}
    
    \textbf{Final Answer:} \textcolor{failFrame}{2} (Correct number, but low confidence / guessing)
    \end{tcolorbox}
    \vspace{0.2cm}

    % --- 正确案例 (GUARD) ---
    \begin{tcolorbox}[
        enhanced,
        title={\textbf{\textcolor{white}{GUARD (Success): Structural Proof Injection}}},
        colframe=succFrame, colback=succBack,
        coltitle=white,
        attach boxed title to top left={yshift=-2mm, xshift=2mm},
        boxed title style={boxrule=0pt, colframe=succFrame},
        boxrule=0.5pt,
        sharp corners=south
    ]
    \small
    \textbf{Reasoning Trace:}\\
    \textit{[...Initial checks for $a=2,3$ match Base Model...]}\\
    For $a=2$ and $a=3$, we find $k=2$. The chain breaks at the 3rd term ($y_3$) or 2nd term ($y_2$).
    I need to determine if $k=3$ is possible for \textit{any} $a$.
    
    \texttt{\textbackslash n\textbackslash n} \txtspiral{Let me assume $k=3$...} \hfill \textbf{\textit{\textcolor{gray}{[$\uparrow$ Entropy Spike Detected]}}}
    
    \vspace{0.1cm}
    \textbf{\textcolor{succFrame}{$\Rightarrow$ GUARD Triggered: Branch 3 (Counterfactual)) "Let me reconsider:"}}
    
    \txtcorrect{Consider modulo 8. Let's pause the search and analyze the structure modulo 8. This might reveal a divisibility property.}
    
    \textbf{Step 1: Parity Check.}
    If $y_1=2^a-1$ is prime, $a$ must be prime.
    If $a=2$ (even prime), we verified $k=2$.
    Assume $a$ is an \textit{odd} prime.
    
    \textbf{Step 2: Modulo 8 Analysis.}
    Sequence: $x_1=a$, $x_2=2a+1$, $x_3=4a+3$.
    Since $a$ is odd, $4a \equiv 4 \pmod 8$.
    Thus, $x_3 = 4a+3 \equiv 4+3 \equiv 7 \pmod 8$.
    
    \textbf{Step 3: Quadratic Residue Argument.}
    Since $x_3 \equiv 7 \pmod 8$, we know that 2 is a \textbf{quadratic residue} modulo $x_3$.
    $\Rightarrow \exists s$ such that $s^2 \equiv 2 \pmod{x_3}$.
    
    \textbf{Step 4: Fermat's Little Theorem.}
    $2^{(x_3-1)/2} \equiv (s^2)^{(x_3-1)/2} \equiv s^{x_3-1} \equiv 1 \pmod{x_3}$.
    Note that the exponent $\frac{x_3-1}{2} = \frac{4a+2}{2} = 2a+1 = x_2$.
    So, $2^{x_2} \equiv 1 \pmod{x_3}$.
    
    \textbf{Step 5: Conclusion.}
    $2^{x_2} - 1 \equiv 0 \pmod{x_3}$.
    This means $x_3$ divides $y_2$. Since $y_2 > x_3$, $y_2$ must be composite.
    Therefore, for any odd prime $a$, the sequence breaks at $y_2$ (or earlier).
    The maximum length is $k=2$ (achieved by $a=2$).
    
    \textbf{Final Answer:} \txtcorrect{2}
    \end{tcolorbox}

    % --- 分析 ---
    \vspace{0.2cm}
    \noindent\textbf{Analysis:} The Base model falls into an "Empirical Spiral," trying to solve a structural number theory problem by brute-force enumeration. Lacking a proof strategy, it wanders through large candidates ($a=89, 127\dots$) hoping to get lucky. GUARD detects the high entropy associated with this aimless search and intervenes with a Counterfactual Branch. By injecting the specific heuristic to "consider modulo 8," it guides the model to construct a rigorous proof using quadratic residues, transforming a guess into a mathematical certainty.

\end{tcolorbox}
\caption{\textbf{Qualitative Comparison on OlympiadBench.} The Base model (Top) attempts to solve the problem by exhaustively checking prime candidates, eventually getting stuck in an infinite search space. GUARD (Bottom) detects the lack of convergence and steers the reasoning towards a structural proof using modular arithmetic and quadratic residues, proving that $k=3$ is impossible.}
\label{fig:case_study_olympiad}
\end{figure*}

\end{document}